%% file: aaai2021.tex
\def\year{2021}\relax
\documentclass[letterpaper]{article} 
\usepackage{aaai21}  
\usepackage{times}  
\usepackage{helvet} 
\usepackage{courier}  
\usepackage[hyphens]{url}  
\usepackage{graphicx} 
\usepackage{booktabs}
\usepackage[monochrome]{color} 
\usepackage{multirow}
\usepackage{makecell}
\usepackage{pdflscape}
\usepackage[table]{xcolor}
\usepackage{natbib}  
\usepackage{caption} 
\usepackage{diagbox}

\definecolor{aquamarine}{rgb}{0.5, 1.0, 0.83}
\definecolor{applegreen}{rgb}{0.0, 0.5, 0.0}
\definecolor{myred}{rgb}{0.8, 0.0, 0.0}
\definecolor{darkgreen}{rgb}{0.0, 0.4, 0.13}
\usepackage{xspace}
\usepackage{comment}

\usepackage[nameinlink,capitalise]{cleveref}

\usepackage{xcolor}
\usepackage[switch]{lineno}

\newcommand\jena[1]{{\color{purple}[#1] - Jena}}

\newcommand\atomic{\textsc{Atomic}\xspace}
\newcommand\atomicTT{\textsc{Atomic$^{20}_{20}$}\xspace}
\newcommand\transomcs{\textsc{TransOMCS}\xspace}
\newcommand\conceptnet{\textsc{ConceptNet}\xspace}

\newcommand\gpttt{\textsc{GPT-3}\xspace}
\newcommand\gptxl{\textsc{GPT2-XL}\xspace}
\newcommand\cometgptxl{\textsc{COMET(GPT2-XL)}\xspace}
\newcommand\cometbart{\textsc{COMET(BART)}\xspace}
\newcommand\comet{\textsc{COMET}\xspace}

\newcommand\CapableOf{\texttt{CapableOf}}
\newcommand\UsedFor{\texttt{UsedFor}}
\newcommand\HasProperty{\texttt{HasProperty}}
\newcommand\AtLocation{\texttt{AtLocation}}
\newcommand\HasA{\texttt{HasA}}

\newcommand\InstanceOf{\texttt{InstanceOf}}
\newcommand\PartOf{\texttt{PartOf}}

\newcommand\MadeOf{\texttt{MadeOf}}

\newcommand\Causes{\texttt{Causes}}

\newcommand\HasSubEvent{\texttt{HasSubEvent}}
\newcommand\MotivatedByGoal{\texttt{MotivatedByGoal}}

\newcommand\Desires{\texttt{Desires}}
\newcommand\NotDesires{\texttt{NotDesires}}
\newcommand\NotCapableOf{\texttt{NotCapableOf}}
\newcommand\NotHasProperty{\texttt{NotHasProperty}}

\newcommand\LocatedNear{\texttt{LocatedNear}}
\newcommand\IsA{\texttt{IsA}}
\newcommand\IsBefore{\texttt{isBefore}}
\newcommand\IsAfter{\texttt{isAfter}}
\newcommand\HinderedBy{\texttt{HinderedBy}}
\newcommand\ObjectUse{\texttt{ObjectUse}}
\newcommand\RelatedTo{\texttt{RelatedTo}}
\newcommand\DistinctFrom{\texttt{DistinctFrom}}
\newcommand\xNeed{\texttt{xNeed}}
\newcommand\xAttr{\texttt{xAttr}}
\newcommand\xEffect{\texttt{xEffect}}
\newcommand\xReact{\texttt{xReact}}
\newcommand\xWant{\texttt{xWant}}
\newcommand\xIntent{\texttt{xIntent}}
\newcommand\oEffect{\texttt{oEffect}}
\newcommand\oReact{\texttt{oReact}}
\newcommand\oWant{\texttt{oWant}}
\newcommand\MadeUpOf{\texttt{MadeUpOf}}
\newcommand\xReason{\texttt{xReason}}
\newcommand\isFilledBy{\texttt{isFilledBy}}

\title{
(\textsc{Comet}-)\textsc{Atomic}${^{20}_{20}}$:\\
On Symbolic and Neural Commonsense Knowledge Graphs
}

\author{
Jena D. Hwang\textsuperscript{\rm 1}\thanks{The authors contributed equally to this work.}, \,
Chandra Bhagavatula\textsuperscript{\rm 1}\footnotemark[1], \,
Ronan Le Bras\textsuperscript{\rm 1}, \,
Jeff Da\textsuperscript{\rm 1}, \,
Keisuke Sakaguchi\textsuperscript{\rm 1},\\
Antoine Bosselut\textsuperscript{\rm 1}\textsuperscript{\rm 3}\, {\normalfont and} \,
Yejin Choi\textsuperscript{\rm 1}\textsuperscript{\rm 2}
}
\affiliations {
    \\
    \textsuperscript{\rm 1} Allen Institute for AI, WA, USA\\
    \textsuperscript{\rm 2} Paul G. Allen School of Computer Science \& Engineering, WA, USA \\
    \textsuperscript{\rm 3} Stanford University, CA, USA \\
    \{jenah, chandrab, ronanl, jeffd, keisukes, antoineb, yejinc\}@allenai.org\\
}

\begin{document}
\maketitle

\begin{abstract}

Recent years have brought about a renewed interest in commonsense representation and reasoning in the field of natural language understanding. The development of new commonsense knowledge graphs (CSKG) has been central to these advances as their diverse facts can be used and referenced by machine learning models for tackling new and challenging tasks. At the same time, there remain questions about the quality and coverage of these resources due to the massive scale required to comprehensively encompass general commonsense knowledge.

In this work, we posit that manually constructed CSKGs will never achieve the coverage necessary to be applicable in all situations encountered by NLP agents. Therefore, we propose a new evaluation framework for testing the utility of KGs based on how effectively implicit knowledge representations can be learned from them.
 
With this new goal, we propose \atomicTT{}, a new CSKG of general-purpose commonsense knowledge containing knowledge that is not readily available in pretrained language models. We evaluate its properties in comparison with other leading CSKGs, performing the first large-scale pairwise study of commonsense knowledge resources. Next, we show that \atomicTT{} is better suited for training \textit{knowledge models} that can generate accurate, representative knowledge for new, unseen entities and events. Finally, through human evaluation, we show that the few-shot performance of GPT-3 (175B parameters), while impressive, remains $\sim$12 absolute points lower than a BART-based knowledge model trained on \atomicTT{} despite using  over 430x fewer parameters.



\end{abstract}

\begin{figure}[t]
\centering
  \includegraphics[width=1\columnwidth]{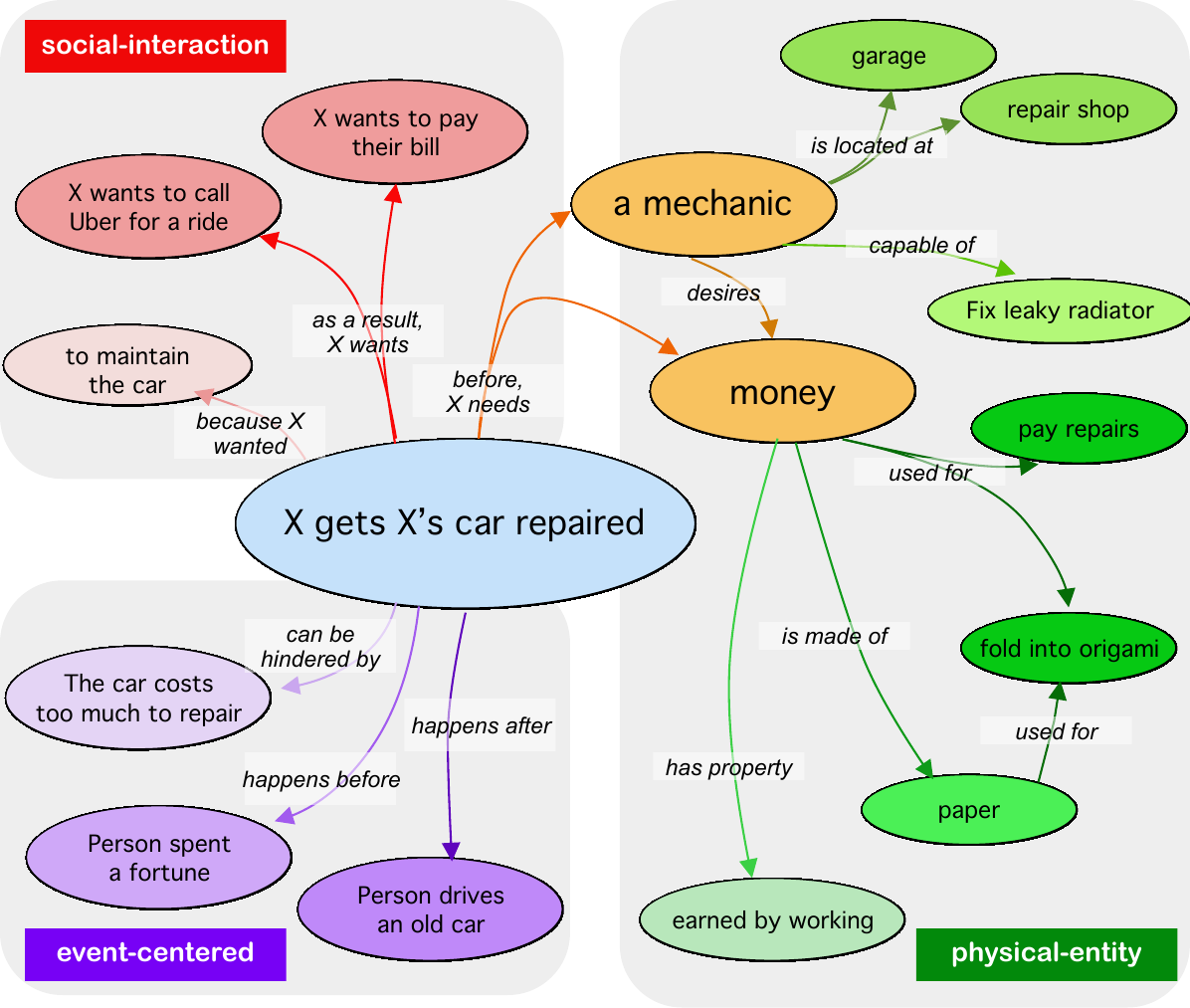}
  \caption{A tiny subset of \atomicTT{}, a large atlas of social and physical commonsense relations. Relations in the top-left quadrant reflects relations from \atomic{}.\footnotemark} 
  
  \label{fig:large_example}
\end{figure}
\footnotetext{\atomicTT{} and \textsc{Atomic}-$2020$ can be used interchangeably, but for brevity we use \atomicTT{} in this paper.}

\begin{figure*}[t]
\centering
\includegraphics[clip,width=0.80\textwidth]{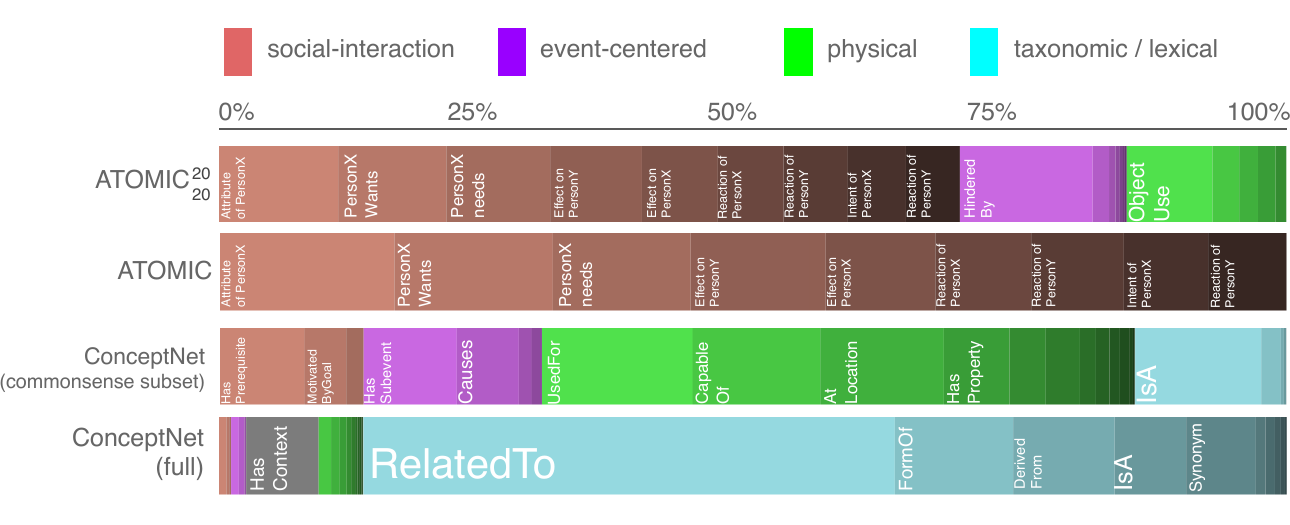}
\caption{\atomicTT{} tuple count distribution compared to \atomic~\cite{sap2018atomic} and \conceptnet, either its commonsense subset~\cite{li-16} or the full set~\cite{speer2017conceptnet}.}
\label{fig:tuple_distribution}
\end{figure*}

\section{Introduction}
\label{sec:introduction}




Commonsense understanding
and reasoning remain long-standing challenges in general artificial intelligence. 
However, large-scale language models have brought tremendous progress in the sub-field of natural language processing. 
Such large-scale language models \citep{openaigpt,bert,brown2020language}  trained on extreme-scale data have been shown to effectively adapt to diverse downstream tasks, achieving significant performance gains across natural language benchmarks \citep{Wang2019SuperGLUEAS}. 
Interestingly, as these models have grown larger (and trained on larger amounts of data), their benchmark performance has continued to improve \cite{Raffel2019ExploringTL} despite limited conceptual improvements, 
leaving open questions regarding 
the source of these remarkable generalization properties. 

Recent work has hypothesized that many of these performance gains could be a result of language models being able to memorize facts in their parameters during training \cite{roberts2020knowledge} that can be leveraged at evaluation time. As a result, a new paradigm of language models as knowledge bases has emerged \cite{Petroni2019LanguageMA}. In this setting, language models are prompted with natural language prefixes or questions, and they express knowledge through language generation. The initial success of this paradigm for representing commonsense knowledge \cite{davison-etal-2019-commonsense,tamborrino-etal-2020-pre}
 has led to the optimistic claim that language models comprehensively encode commonsense knowledge, and remove the need for structured knowledge resources. 

We take a more skeptical view of this capacity of language models -- \textit{Does scaling up language models actually endow them with commonsense knowledge?} While language models can successfully express certain types of knowledge, their best results are observed in narrowly specific conditions -- we show (cf. \S\ref{sec:generalization}) that they perform better when evaluated on knowledge bases that prioritize ontological relations and whose examples resemble language-like assertions (e.g., mango \IsA{} fruit).\footnote{An observation supported by \citet{brown2020language}'s \gpttt{} model, whose best few-shot performance on commonsense knowledge benchmarks comes on the PhysicalIQA \cite{Bisk2020Piqa} and HellaSwag \cite{zellers-etal-2019-hellaswag} datasets.} Consequently, the types of knowledge that can be directly accessed through the language model's interface remains limited. 

However, prior work has also shown that training language models on knowledge graph tuples leads them to learn to express their implicit knowledge directly \citep{Bosselut2019COMETCT}, allowing them to provide commonsense knowledge on-demand. These adapted \textit{knowledge models} have exhibited promising results on commonsense benchmarks compared with methods that require linking entities to knowledge graphs \cite{shwartz2020unsupervised,liu-etal-2020-commonsense}. Inspired by these successes, we propose a dual use for commonsense knowledge bases going forward: as static graphs that can be linked to for discrete knowledge access, and as resources for adapting language models to hypothesize commonsense knowledge about un-annotated entities and events.


With this second purpose in mind, we propose evaluating commonsense knowledge resources based on the complementary information they can bring to pretrained language models.
We construct \atomicTT{}, a new, high-quality knowledge graph with $1.33$M commonsense knowledge tuples across $23$ commonsense relations.
We compare \atomicTT{} with respect to its coverage and accuracy in competition with other highly used CSKGs, such as \conceptnet~\citep{speer2017conceptnet}. Our results show that \atomicTT{} is able to cover more correct facts about more diverse types of commonsense knowledge than any existing, publicly-available commonsense knowledge resource.
However, our results also indicate that there remains a large amount of exclusivity between these KGs, highlighting the challenge of creating resources that cover the scale and diversity of general commonsense knowledge.
 

Furthermore, we formalize the \comet framework of \citet{Bosselut2019COMETCT} across different seed language models and training knowledge graphs, and evaluate the commonsense knowledge hypothesized by these adapted \textit{knowledge models}. 
Our empirical study yields two promising conclusions. First, it confirms that KG-adapted language models learn to express knowledge more precisely than naive language models trained only on language. And second, we show that \atomicTT{} as a transfer resource leads to \comet models that achieve the largest increase over their seed language model (across all seed LMs) for the commonsense knowledge types it covers, validating the importance of constructing knowledge resources with examples of knowledge not readily found in language models.
 

\textbf{Key Contributions:} 
In summary, we make three key contributions in this paper. We present \atomicTT{}---a new commonsense knowledge graph covering social, physical, and eventive aspects of everyday inferential knowledge (cf. \S\ref{sec:atomicTT}). Next, we compare \atomicTT{} with other prominent CSKBs head-to-head and show that our new \textit{symbolic} knowledge graph is more accurate than any current CSKB (see Table \ref{tab:precision-results}) (cf. \S\ref{sec:kb-comparison-results}). Finally, we show that our new \textit{neural} knowledge model \comet{}-\atomicTT{} successfully transfers \atomicTT{}'s declarative knowledge to beat \gpttt{}, the largest pre-trained language model, in spite of using ~400x fewer parameters (see Table \ref{tab:human-eval-generations}) (cf. \S\ref{sec:generalization}). This demonstrates the utility and importance of high-quality symbolic knowledge provided by \atomicTT{} to generalize on commonsense information that LMs cannot expressively capture on their own (cf. \S\ref{sec:discussion}).



\section{Background}
\label{sec:background}
\paragraph{Commonsense Knowledge Graphs}
Large scale commonsense knowledge graphs are ubiquitous tools in natural language processing tasks as access to their facts allows models to learn to reason over commonsense knowledge to make predictions \citep{Lin2019KagNetKG,feng2020scalable}. In this work, we evaluate three existing knowledge graphs, \conceptnet, \atomic, and \transomcs on their coverage and precision relative to our new resource \atomicTT.\footnote{We were unable to include Cyc \cite{lenat1995cyc} in our study due to the discontinuation of its research license and the cost of the commercial license (over $\$1$M). \conceptnet includes a subset of Cyc -- OpenCyc.} 

The \textbf{ \conceptnet} (v5.7) knowledge graph \cite{speer2017conceptnet} consists of 36 relations focusing mostly on taxonomic and lexical knowledge (e.g., \texttt{RelatedTo}, \texttt{Synonym}, \texttt{IsA}) and physical commonsense knowledge (e.g., \MadeOf, \PartOf). 
\conceptnet(v5.7) contains 3.4M entity-relation tuples (in English) collected by crowdsourcing and merged with existing knowledge databases from DBPedia, WordNet, Wiktionary, and OpenCyc.
Since the knowledge are derived from human efforts, the accuracy of \conceptnet(v5.7) knowledge is fairly high, though the quality does vary depending on the sources of knowledge and relation types. 
However, as highlighted in~\cite{davis-marcus2015,sap2018atomic}, and shown in Figure~\ref{fig:tuple_distribution}, the coverage of \conceptnet(v5.7) is limited to mostly taxonomic, lexical, and object-centric physical commonsense knowledge.
In fact, out of 3.4M tuples, 90\% of them correspond to taxonomic (e.g., \texttt{IsA}) or lexical (e.g., \texttt{Synonym}, \texttt{RelatedTo}) knowledge, making the commonsense portion of \conceptnet(v5.7) relatively small. 

The \textbf{\atomic}~\cite{sap2018atomic} knowledge graph consists of 880K of tuples across 9 relations that cover social commonsense knowledge (e.g, X gets X's car repaired \texttt{xIntent} to maintain the car), including dynamic aspects of events such as causes and effects, \texttt{if-then} conditional statements, and mental states.
The \atomic dataset is collected and validated completely through crowdsourcing. 

The \textbf{\transomcs}~\cite{zhang2020TransOMCS} knowledge graph consists of 18.48M tuples that were automatically converted from syntactic parses of sentences from various web sources including Wikipedia, Yelp, 
and Reddit.
The set of relations used for the mapping is copied from \conceptnet. 
Although \transomcs is much larger than other commonsense knowledge graphs, the precision of the extracted knowledge is significantly lower compared to other resources (cf. \S\ref{sec:kb-comparison-precision}), and performs poorly as an adaptation resource relative to other KGs (cf. \S\ref{sec:generalization}).

For this work we have selected three large scale CSKGs that retain a closed class of relational types that are comparable to one another. Other commonsense KBs in existence such as Quasimodo \cite{romero2019commonsense} provide a wider variety of fine-grained relations.

\vspace{2mm} \noindent \textbf{Language Models as Knowledge Bases}
Recent work hypothesizes that pretrained language models represent commonsense knowledge implicitly \citep{Petroni2019LanguageMA,roberts2020knowledge}.
However, the results motivating these observations are often limited to narrowly scoped subsets of commonsense knowledge that primarily include taxonomic knowledge (e.g., mango \IsA{} fruit) and that are often found explicitly stated in text. However, commonsense facts are often implied \citep{gordon2013reporting}, and as will be seen in our studies (cf. \S\ref{sec:kb-comparison-results}), state of the art neural models struggle to express implicit commonsense knowledge that involves complex relationships. 

To overcome this limitation, \citet{Bosselut2019COMETCT} take the best of both worlds between commonsense knowledge graphs and pretrained language models.
The commonsense transformer, or \comet, adapts pretrained neural language models by training on example tuples from commonsense knowledge graphs.
It takes a head/source phrase and a relation (e.g., take a nap \Causes{}) and generates the tail/target phrase (e.g., have energy).
\citet{Bosselut2019COMETCT} show that \comet trained on the \conceptnet and \atomic knowledge graphs is able to adapt to generate novel (and valid) commonsense knowledge tuples.

Importantly, these neural \emph{knowledge models} can produce commonsense knowledge on-demand for any head entity that can be expressed through language. This flexibility allows them to be used out-of-the-box, and they have been applied to new, previously unexplored tasks, such as sarcastic comment generation~\cite{chakrabarty-etal-2020-r}, therapy chatbots~\cite{Kearns2020AWI}, and automated story plot generation~\cite{ammanabrolu2020automated}. These contributions show that progress on knowledge models opens up new downstream applications that were challenging to model before.

\section{\atomicTT}
\label{sec:atomicTT}

We present \atomicTT{}, a commonsense knowledge graph with $1.33$M everyday inferential knowledge tuples about entities and events. \atomicTT{} represents a large-scale commonsense repository of textual descriptions that encode both the social and the physical aspects of common human everyday experiences, collected with the aim of being complementary to commonsense knowledge encoded in current language models. \atomicTT{} introduces $23$ commonsense relations types. They can be broadly classified into three categorical types: $9$ commonsense relations of social-interaction, $7$ physical-entity commonsense relations, and $7$ event-centered commonsense relations concerning situations surrounding a given event of interest. The full inventory of \atomicTT{} relations is listed in Table~\ref{tab:atomic2020-newrelations}.

In terms of physical and event-centered commonsense, by far, the two largest new relations in \atomicTT{} are \ObjectUse{} and \HinderedBy{}. For \ObjectUse{}, we focused on \textit{affordances} of everyday objects such as ``popcorn bucket'' that may be used for ``holding popocorn'' or ``storing things''. For \HinderedBy{}, we explore the notion that many events in real world can be defeasible~\cite{lascarides1991discourse} by collecting hindrances to goals that may be useful for tasks such as counterfactual reasoning. For example X's desires to adopt a cat may be hindered by finding out that X is allergic to cats, which would necessitate X to adjust future actions accordingly (say, opt for hypoallergenic options like tortoises). 

\input{tables/tab-atomic2020-all-relations}

In the case of \ObjectUse{}, we collected over 130K everyday object-use pairs by asking crowdworkers for necessary objects and their uses for each event in \atomicTT{}. For example, given ``X eats popcorn'' we elicited items such as ``popcorn bucket'' with their various expected uses. The number also reflects \textit{atypical} usages gathered in a separate pass where workers were asked to provide creative or resourceful but \textit{feasible} uses of the objects. Given ``popcorn bucket'', for instance, one might ``wear it as a hat'' for, say, a costume party.  For \HinderedBy{}, we crowdsourced over 100K tuples of hindrances to existing \atomicTT{} events, asking the workers to provide situations or events that might pose as deterrence should the event be considered an achievable goal (see Appendix~\ref{appendix:atomic2020-details} 
for further details). 
%
For social-interaction commonsense, we primarily incorporated tuples from \atomic{}, 
but also crowdsourced an additional 34K tuples using the same approach as \citet{sap2018atomic}.

\atomicTT{} also pulls commonsense tuples from the English subset of \conceptnet{}(v5.7) (latest version available; \citealt{speer2017conceptnet}).\footnote{A \conceptnet{}(v5.7) fact is considered English if both the head and tail concepts are marked with `/en/' in the edge id.} Of the $3.4$M English tuples in \conceptnet{}(v5.7), a small subset of 172K tuples was selectively chosen to be integrated into \atomicTT{} via elimination and crowdsourcing. This subset represents data carefully identified to reflect commonsense information dealing with qualitative human experiences. Among the eliminated data are tuples with edge weight $\leq 0.5$,  dictionary or etymologically based knowledge (e.g., synonyms/antonyms, inflections), lexical hyper/hyponymic lexical relationships such as \IsA{} or \InstanceOf{}, and relations based on lexical co-occurrence (e.g., \RelatedTo{} or \LocatedNear{}), which are easily recoverable from language models.\footnote{\conceptnet{} 5.7 defines weight as ``the strength with which this edge expresses this assertion''. A pilot crowdsource assessment step found any tuple with weight $\leq 0.5$ unreliable w.r.t. its validity.} After selective removal of these relations and a post-processing step to ensure the removal of deterministic information such as geographic facts (e.g., ``shenzhen'' \AtLocation ``china''), tuples from each \conceptnet{} were examined for further splits or joins to align with the existing structure of \atomicTT{}. A random 10\% tuples from each selected relations were then put through crowdsourced validity testing (akin to the process described later in \S\ref{sec:kb-comparison-precision}). Tuples that were directly incorporated without further edits passed with an acceptance rate of 93\% or higher. A subset of relations (i.e., \CapableOf, \HasProperty, \MotivatedByGoal) were put through additional crowdsourcing to weed out tuples that were either invalid or found to hold prejudiced descriptions of human entities. In the end, only 5 relations (marked with an asterisk in Table \ref{tab:atomic2020-newrelations}) retain the \conceptnet{}'s original meaning with a few relations that are cognates in \atomicTT{} (more details in Appendix~\ref{appendix:atomic2020-details}).

\section{Symbolic Knowledge Graph Comparison}
\label{sec:kb-comparison-results}
\input{sections/kg-comp-symbolic}

\section{Neural Knowledge Graph Comparison}
\label{sec:generalization}
\input{sections/kg-comp-neural}

\section{Discussion}
\label{sec:discussion}

\paragraph{Do pretrained language models already encode commonsense knowledge?} Our conclusions on this subject are mixed and hinge on the ambiguous meaning of what it means to \textit{encode} knowledge. Despite the conclusions of prior work \citep{Petroni2019LanguageMA,roberts2020knowledge,tamborrino-etal-2020-pre}, our results in Table~\ref{tab:human-eval-generations} are clear that language models fail to express large varieties of knowledge when prompted for it in a zero-shot manner. When converted to \comet models by training on a knowledge graph, their performance at hypothesizing knowledge tuples skyrockets -- 47.9\% absolute difference between \cometbart{} and \gptxl{} on \atomicTT. 

However, the evaluation tuples are adversarially selected to not include head entities that were in the training set. The model must generalize its learned representations of relations to entities it has not observed these relationships for 
during fine-tuning, meaning the representation of these entities is solely formulated from learning language. As a result, language models may still \textit{encode} this knowledge in their parameters, even if they are not capable of \textit{expressing} it directly. With this framing in mind, the COMET training paradigm proposed by \citet{Bosselut2019COMETCT} can perhaps be viewed less as a means of learning \textit{knowledge} from KGs, and more as a method of learning an \textit{interface} for language models to hypothesize encoded knowledge through language generation. We look forward to future work in this space that attempts to disentangle these two ideas.

\vspace{2mm} \noindent \textbf{What considerations should be made when designing commonsense knowledge resources?} 
Based on our results in Section~\ref{sec:generalization}, we outline desiderata for the design and development of future commonsense knowledge graphs. Because certain types of knowledge are already encoded and expressible by pretrained language models, CSKG designers should focus on collecting examples and categories of knowledge that are less likely to be known by language models. For example, of the 378 test tuples evaluated by the \gptxl{} zero-shot model that contained the \HinderedBy{} relation, only 1.3\% were deemed plausible by human raters -- jumping to 85\% plausibility for \cometbart{} -- pointing to an advantage in constructing \atomicTT{} with this relationship in mind (see Appendix~\ref{app:neural} 
for per-relation accuracy).

Second, commonsense knowledge resources should be designed with the goal of accuracy and relationship coverage. Because language models exhibit powerful adaptation \citep{brown2020language}, they can generalize many commonsense relationships as long they have examples on which to train. Consequently, we should construct commonsense resources that encapsulate larger numbers of relations so the knowledge in pretrained language models can be grounded to a variety of relationships. However, language models also benefit from learning from precise examples. Being able to train on a large collection of examples from \transomcs (see Appendix~\ref{app:neural})
did not allow \comet models to generalize to unseen entities as these examples were not of sufficient quality (See Table~\ref{tab:precision-results}). Resources should be carefully validated for the quality of their facts, an example set by \citet{speer2017conceptnet} and \citet{sap2018atomic}.

\section{Conclusion}

In this work, we formalize a use for commonsense knowledge graphs as transfer learning tools for pretrained language models. With this new purpose, we hypothesize that commonsense knowledge graphs should be designed to contain knowledge that is not already expressible by language models without difficulty (e.g., not taxonomic and lexical knowledge). Consequently, we propose \atomicTT, a novel commonsense knowledge graph containing tuples whose relations are specifically selected to be challenging for pretrained language models to express. Our empirical studies demonstrate that \atomicTT contains high-accuracy knowledge tuples across multiple novel relations not found in existing CSKGs or expressible by LMs. Furthermore, we show that \atomicTT can be effectively used as a training set for adapting language models as \textit{knowledge models} to generate high quality tuples on-demand.

\section*{Acknowledgements}
\input{sections/acknowledgements}

\bibliography{aaai2021}

\clearpage
\appendix
\input{sections/appendix}

\end{document}

%% file: tables/tab-atomic2020-all-relations.tex
\begin{table}[t]
    \small
    \centering
    \setlength\tabcolsep{3pt}
    \begin{tabular}{p{1em}p{4.5em}p{5.5em}p{9.5em}r}
    \toprule
    &\textbf{Head}  & \textbf{Relation} & \textbf{Tail} & \textbf{Size}\\
    \midrule
    \multirow{10}{*}{\rotatebox[origin=rb]{90}{{\textsc{Physical-Entity}}}} & \multirow{5}{*}{bread} & ObjectUse & make french toast & $165{,}590$\\ 
    \cmidrule{3-5}
    & & AtLocation$^*$ & basket; pantry & $20{,}221$\\
    \cmidrule{3-5}
    & & MadeUpOf & dough; wheat  & $3{,}345$\\
    \cmidrule{3-5}
    & & HasProperty$^*$ & cooked; nice to eat & $5{,}617$\\
    \cmidrule{2-5}
    &\multirow{4}{*}{baker} & CapableOf$^*$ & coat cake with icing & $7{,}968$\\
    \cmidrule{3-5}
    & & Desires$^*$ & quality ingredients & $2{,}737$\\
    \cmidrule{3-5}
    & & Not Desires$^*$ & bad yeast & $2{,}838$\\
    \midrule
    \midrule
    \multirow{10}{*}{\rotatebox[origin=rb]{90}{{\textsc{Event-Centered}}}}  & \multirow{8}{*}{\makecell{X runs out\\of steam}} & IsAfter & X exercises in the gym & $22{,}453$\\
    \cmidrule{3-5}
    & & HasSubEvent & become tired  & $12{,}845$\\
    \cmidrule{3-5}
    & & IsBefore & X hits the showers & $23{,}208$\\
    \cmidrule{3-5}
    & & HinderedBy & drinks too much coffee & $106{,}658$\\
    \cmidrule{3-5}
    & & Causes & takes a break & $376$\\
    \cmidrule{3-5}
    & & xReason & did not eat breakfast & $334$ \\
    \cmidrule{2-5}
    & \makecell{X watches\\\_\_\_ anyway} & isFilledBy & the game; the TV & $33{,}266$\\
    \midrule
    \midrule

    
    \multirow{12}{*}{\rotatebox[origin=rb]{90}{{\textsc{Social-Interaction}}}}  & \multirow{7}{*}{\makecell{X runs out\\of steam}} & xNeed & do something tiring & $128{,}955$ \\
    \cmidrule{3-5}
    & & xAttr & old; lazy; lethargic  & $148{,}194$\\
    \cmidrule{3-5}
    & & xEffect & drinks some water & $115{,}124$\\
    \cmidrule{3-5}
    & & xReact & tired & $81{,}397$\\
    \cmidrule{3-5}
    & & xWant & to get some energy & $135{,}360$ \\
    \cmidrule{2-5}
    & \multirow{5}{*}{\makecell{X votes\\for Y}} & xIntent & to give support & $72{,}677$\\
    \cmidrule{3-5}
    & & oEffect & receives praise & $80{,}166$\\
    \cmidrule{3-5}
    & & oReact & grateful; confident  & $67{,}236$\\
    \cmidrule{3-5}
    & & oWant & thank X; celebrate   & $94{,}548$\\
    \bottomrule
    \end{tabular}
    \caption{Relations in \atomicTT{} along with illustrative examples and their respective size. Relations that reflect semantically identical categories to \conceptnet{} is marked with an asterisk ($^*$).}
    \label{tab:atomic2020-newrelations}
\end{table}

%% file: sections/kg-comp-symbolic.tex


In this work, we compare our new \atomicTT{} knowledge graph to three other prominent CSKGs: \atomic{} \cite{sap2018atomic}, \conceptnet{}\footnote{Hereafter, as we focus on CSKGs, by ConceptNet, we refer to the commonsense subset, unless specified otherwise.}~\cite{li-16}, and \transomcs~\cite{zhang2020TransOMCS}. We measure the accuracy of tuples in each KG and compare the coverage of each CSKG w.r.t. other CSKGs head-to-head.

\subsection{Accuracy Assessment}
\label{sec:kb-comparison-precision}

\input{sections/kb-comparison-precision}

\subsection{Coverage Assessment}
\label{sec:kb-comparison-coverage}
\input{sections/kb-comparison-coverage}

\input{sections/kb-comparison-results}

\begin{table}[t]
\small
\setlength\tabcolsep{3pt} 
\begin{tabular}{llrrr}
\toprule
 \textbf{KG}           &   \textbf{Model}             & \textbf{Accept} & \textbf{Reject} & \textbf{\makecell{No\\Jdgm.}} \\
\midrule            

 \multirow{4}{*}{\atomicTT{}} 
 & \gptxl{}        & 36.6                    & 62.5                   & 0.9                           \\
   & \gpttt{} & 73.0 & 24.6 & 2.5       \\
 & \cometgptxl{} & 72.5                    & 26.6                   & 0.9                           \\
            & \cometbart{}    & \textbf{84.5}                    & \textbf{13.8}                   & 1.7                           \\
\midrule
      & \gptxl{}        & 38.3                    & 61.2                   & 0.4                           \\
\atomic{}            & \cometgptxl{} & 64.1                    & 34.7                   & 1.2                           \\
            & \cometbart{}    & \textbf{83.1}                    & \textbf{15.3}                   & 1.6                           \\
\midrule
  & \gptxl{}        & 50.3                    & 42.1                   & 7.7                           \\
\conceptnet{}            & \cometgptxl{} & 74.5                    & 19.0                   & 6.4                           \\
            & \cometbart{}    & \textbf{75.5}                    & \textbf{17.9}                   & 6.6                           \\
\midrule
   & \gptxl{}        & \textbf{28.7}                    & \textbf{53.5}                   & 17.8                           \\
 \transomcs{}           & \cometgptxl{} & 26.9                    & 60.9                   & 12.2                           \\
            & \cometbart{}    & 23.8                    & 65.9                   & 10.3                  \\
            
\bottomrule
\end{tabular}
\caption{Human evaluation of generation accuracy ($\%$). Each model uses greedy decoding to generate the \emph{tail} of 5K randomly-sampled test prefixes (\emph{head}, \emph{relation}) from each knowledge graph. \gptxl{}, \gpttt{} and BART have 1.5B, 175B and 440M parameters, respectively.}
\label{tab:human-eval-generations}
\end{table}

%% file: sections/kb-comparison-precision.tex
In order to assess the accuracy of the knowledge represented, 3K random instances were extracted from each of the knowledge graphs for a crowdsourced evaluation of the tuples.

\begin{table}[t]
\center
\small
\begin{tabular}{lrrr}
\toprule
\textbf{Knowledge Base} & \textbf{Accept}	& \textbf{Reject}	&\textbf{No Judgment}\\
\midrule
$\atomicTT$   & \textbf{91.3} & \textbf{6.5} & 2.2\\
$\atomic$     & 88.5 & 10.0 & 1.5\\
$\conceptnet$ & 88.6 & 7.5 & 3.9\\
$\transomcs$  & 41.7 & 53.4 & 4.9\\
\bottomrule
\end{tabular}
\caption{Accuracy - Percentage ($\%$) of tuples in the knowledge base evaluated by human crowdworkers as either always true or likely (Accept), farfetched/never or invalid (Reject), or unclear (No Judgment).}
\label{tab:precision-results}
\end{table}

\begin{table}[t]
\small
\setlength\tabcolsep{3pt} 
\begin{tabular}{ccccc}
\toprule
\textbf{\atomicTT{}} & \textbf{\atomic{}} & \textbf{\textit{Relation}} & \textbf{CN} & \textbf{T-OMCS} \\
\midrule                  

\textbf{92.3} &  & AtLocation* & 89.4 & \fbox{34.3} \\
\textbf{93.9} &  & CapableOf* & \fbox{84.4} & \fbox{50.0} \\
\textbf{94.6} &  & Causes & \fbox{90.0} & \fbox{50.0} \\
\textbf{96.9} &  & Desires* & 96.3 & \fbox{48.2} \\
\textbf{93.9} &  & HasProperty* & \fbox{86.3} & \fbox{52.4} \\
82.3 &  & ObjUse/UsedFor & \fbox{\textbf{96.3}} & \fbox{31.6} \\
\textbf{98.5} &  & NotDesires* & 96.3 &  \\
\textbf{96.9} &  & HasSubevent & \fbox{88.1} &  \fbox{57.7} \\
 &  & HasFirstSubevent & \textbf{93.8} & 52.4 \\
 &  & HasLastSubevent & \textbf{95.6} & 38.2 \\
 &  & HasPrerequisite & \textbf{94.4} & 30.0 \\
75.4 &  & MadeUpOf/MadeOf & \fbox{\textbf{88.1}} & \fbox{ 15.9}  \\
 &  & PartOf & \textbf{71.9} & 46.5 \\
 &  & HasA & \textbf{77.5} & 43.5 \\
\textbf{96.9} &  & HinderedBy &  &  \\
\textbf{96.2} &  & isAfter &  &  \\
\textbf{95.4} &  & isBefore &  &  \\
\textbf{96.2} &  & isFilledBy &  &  \\
 &  & ReceiveAction & \textbf{84.4} & 56.4 \\
\textbf{91.5} & 86.3 & oEffect &  &  \\
\textbf{91.5} & 87.7 & oReact &  &  \\
88.5 & \textbf{89.5} & oWant &  &  \\
87.7 & \textbf{91.0} & xAttr &  &  \\
80.8 & \textbf{87.2} & xEffect &  &  \\
\textbf{93.1} & 89.9 & xIntent/MotivByGoal & 84.4 & \fbox{27.1} \\
\textbf{87.7} & 85.1 & xNeed &  &  \\
90.8 & \textbf{91.3} & xReact &  &  \\
\textbf{96.2} &  & xReason &  &  \\
82.3 & 88.4 & xWant/CausesDesire & \textbf{90.0} &  \fbox{35.9} \\

\bottomrule
 \end{tabular}
\caption{KG accuracy values broken down by relation. Boxed cells indicate statistically significant difference from \atomicTT{} values. 
Relational \textit{cognates} have been grouped together and \textit{exact matches} are asterisked (*) (cf. Table~\ref{tab:atomic2020-newrelations}).}
\label{tab:precision:breakdown}
\end{table}

\input{sections/crowdsourcing}

\vspace{2mm} \noindent \textbf{Results.} \atomicTT{} outperforms other KGs in crowdsourced accuracy as shown in Table~\ref{tab:precision-results}.\footnote{Overall inter-rater agreement measured by Fleiss'~$\kappa$ of 0.46 (moderate agreement; \citealt{fleiss1971measuring}).} \atomic{} ties with \conceptnet{} with reasonably high accuracy, while \transomcs{} lags behind others with far lower accuracy. We provide a per-relation breakdown of accuracies in Table~\ref{tab:precision:breakdown}. 

Between \atomicTT{} and \atomic{}, the variations in the assessed accuracies are not found to be statistically significant. Among the \atomicTT{} and \conceptnet{} relations that represent \textit{exact matches} (marked with * in Table~\ref{tab:precision:breakdown}), the differences are either not statistically significant or when they are, \atomicTT{} improves upon the associated facts, reflecting that the preprocessing stages of \conceptnet{} integration were helpful in improving the quality of these relations (\S\ref{sec:atomicTT}).  
Among \textit{cognates} in \atomicTT{} and \conceptnet{} relations, two sets of relations fare significantly worse in \atomicTT{} than in \conceptnet{}. In the case of \ObjectUse{}/\UsedFor{}, this is likely due to the fact that \atomicTT{}'s \ObjectUse{} includes atypical affordances (cf. \S\ref{sec:atomicTT}).
In an annotation setting where workers are asked to evaluate the truth or likelihood of an assertion rather than feasibility of use, a portion of the atypical usages are seen as `farfetched' and thus, rejected. In the case of \MadeUpOf{}/\MadeOf{}, there may be some room for improvement for \atomicTT{}. Unlike the \atomicTT{}'s \texttt{HasSubEvent} label that successfully joins together \conceptnet{}'s \textsc{Has(First/Last)Subevent} labels for an improved accuracy, \atomicTT{}'s \MadeUpOf{} union of \MadeOf{}, \PartOf{}, and a subset of \HasA{}, did not seem to have resulted in improved quality. The rest of the \atomicTT{} cognates see a significantly higher or similar accuracy in comparison to \conceptnet{}.


%% file: sections/crowdsourcing.tex
\vspace{2mm} \noindent \textbf{Qualifying Crowdsource Workers.} The  evaluation was carried out through crowdsourcing on the Amazon Mechanical Turk platform. To ensure high-quality annotations, we qualified a pool of 173 workers through a paid qualification task that tested their ability to follow directions and provide reasonable answers to the qualification test. The qualification test contained 6 manually selected tuples from \atomic and \conceptnet, including both easy and tricky relations to annotate. A worker was qualified if they provided $100$\% acceptable answers. Workers providing $5$ of $6$ correct answers were also accepted only when they provided a reasonable written substantiation for their incorrect choice. Workers were paid an average of \$$15$ per hour for their evaluations. 

\vspace{2mm} \noindent \textbf{Human Evaluation Setup.}  Workers were presented with knowledge tuples in the form of (\textit{head}, \textit{relation}, \textit{tail}) for annotation. To expedite the human assessment of the tuples, each \textit{relation} (e.g., \xWant{} or \AtLocation{}) was translated into a human-friendly natural language form (e.g., ``as a result, PersonX wants'' and ``located or found at/in/on'', respectively; cf. Appendix~\ref{app:evaluation-details}).
The workers were asked to rate the tuples along a 4-point Likert scale: \textit{always/often} -- the knowledge assertion presented is always or often true, \textit{sometimes/likely} -- it is sometimes or likely true, \textit{farfetched/never} -- it is false or farfetched at best, and \textit{invalid} -- it is invalid or makes no sense. Any tuples receiving the former two labels are ranked as \textbf{Accept} and latter two as \textbf{Reject}. The workers were also given a choice to opt out of assessment if the concepts were too unfamiliar for a fair evaluation (\textbf{No Judgment}). Each task (HIT) included 5 tuples of the same relation type, and each tuple was labeled by 3 workers. For the results, we take the majority vote among the 3 workers.

%% file: sections/kb-comparison-coverage.tex
We make a pairwise comparison between the CSKGs to assess their coverage with regards to the commonsense knowledge they contain. 
For a reliable head-to-head comparison, we map relations and tuples between various KGs.

\vspace{2mm} \noindent \textbf{Mapping Relations.} 
Since \atomicTT is built on existing \atomic relations, we primarily need to align relations between $\atomicTT$ and \conceptnet. We manually align them based on the definitions for the labels as supplied by the two graphs, then the resulting alignment was verified by sampling at random approximately 20 instances per relation.

\vspace{2mm} \noindent \textbf{Mapping Tuples.} In order to resolve syntactic differences in how the concepts are expressed in each of the KGs (e.g., $\atomic$'s ``PersonX eats breakfast'' vs. \conceptnet's ``eat breakfast''), we preprocess each of the head and tail concepts of each tuple in each KG in the following manner: (1) the concept is lowercased and stripped of extra spaces, punctuations, and stopwords; (2) any exact tuple duplicates within each KB removed, and (3) remaining content words are lemmatized according to their POS category. For $\atomic$ and $\atomicTT$, an extra step is added to remove mentions of ``PersonX'', ``PersonY'' and ``PersonZ'' if occurring at the beginning of a string, and to replace with `person` if they occur elsewhere (e.g, ``PersonX greets PersonY'').

\vspace{.5em} \noindent \textbf{Metrics.} We use two metrics to evaluate the coverage of knowledge graphs. 
For each pair of CSKGs, we compute precision and recall with respect to a target KG. \textbf{Coverage precision} assesses the proportion of tuples in the source KG that are correct according to tuples in the target KG. \textbf{Coverage recall} reflects the proportion of tuples in the target KB that the tuples in the source KB successfully recalled.



\begin{table}[t]
\small
\begin{tabular}{lrrrr}
\toprule
 & \multicolumn{4}{l}{\textbf{Target KB$\rightarrow{}$}}\\
 \cmidrule{2-5}
\textbf{Source KB$\downarrow{}$}   & \atomic{} & CN& T-OMCS & \atomicTT{} \\
\midrule                  
$\atomic$      & -        & 0.1      & 0.0  & 100.0  \\
$\conceptnet$  & 0.3   & -           & 5.5  & 45.6   \\
$\transomcs$   & 0.0   & 0.4      & -       & 0.3  \\
$\atomicTT$    & 60.2   & 9.3      & 1.4  & -        \\
\bottomrule
\end{tabular}
\caption{Coverage Precision - Average number of times (in $\%$) a tuple in Source KB is found in Target KB.}
\label{tab:coverage:precision}
\end{table}

\begin{table}[t]
\small
\begin{tabular}{lrrrr}
\toprule
 & \multicolumn{4}{l}{\textbf{Target KB$\rightarrow{}$}}\\
 \cmidrule{2-5}
\textbf{Source KB$\downarrow{}$}   & $\atomic$ & CN& T-OMCS & \atomicTT \\
\midrule                  
$\atomic$      & - & 0.3      & 0.0  & 60.1  \\
$\conceptnet$  & 0.1   & -           & 0.3  & 8.9   \\
$\transomcs$   & 0.0   & 7.6      & -     & 1.3  \\
$\atomicTT$    & 100.1$^\dagger$   & 47.8      & 0.4  &   - \\
\bottomrule
\end{tabular}
\caption{Coverage Recall - Average number of times (in $\%$) a tuple in Target KB is found in Source KB. $^\dagger$This value is greater than 100 because multiple tuples in \atomicTT{} can map to the same tuple in \atomic. }
\label{tab:coverage:recall}
\end{table}

%% file: sections/kb-comparison-results.tex
\vspace{2mm} \noindent \textbf{Results.} Tables \ref{tab:coverage:precision} and \ref{tab:coverage:recall} show a pairwise coverage precision and recall assessment among the CSKGs. \atomicTT{} shows the widest coverage:
\atomicTT{} is able to recall all of \atomic{} (as expected) and just under half of \conceptnet{}. There is very little overlap between \atomic{} and \conceptnet{}, which is unsurprising as all of \atomic  knowledge is focused on social behaviors \conceptnet{} does not cover while \conceptnet{} leans on physical commonsense which falls outside \atomic{}'s scope. Overall, \transomcs{} intersects very little with any of the other three KBs.

%% file: sections/kg-comp-neural.tex
Language models are powerful tools for representing knowledge, but their ability to serve as generative knowledge bases is limited by the fact they are directly trained to represent the distribution of language. Previous work shows knowledge graphs can help language models better transfer as knowledge engines \citep{Bosselut2019COMETCT} by re-training them on examples of structured knowledge. As a result, a new purpose for knowledge graphs is to be useful in helping language models generalize to hypothesizing knowledge tuples.

\begin{table*}[t]
  \centering
  \small
  \begin{tabular}{llcccccccc}
\toprule
                &       &   Bleu-1 &   Bleu-2 &   Bleu-3 &   Bleu-4 &   METEOR &   ROUGE-L &   CIDEr &   BERT Score \\
 \midrule
 \multirow{4}{*}{\atomicTT{}} 
& \gptxl{} &    0.101 &    0.028 &    0.010 &    0.003 &    0.082 &     0.098 &   0.047 &        0.395 \\
 & \gpttt{} &    0.299 &    0.153 &    0.081 &    0.048 &    0.182 &     0.255 &   0.175 &        0.540 \\ 
  & \cometgptxl{}     &    0.407 &    0.248 &    0.171 &    0.124 &    0.292 &     0.485 &   0.653 &        0.638 \\
 & \cometbart{}       &    \textbf{0.469} &    \textbf{0.286} &    \textbf{0.189} &    \textbf{0.130} &    \textbf{0.330} &     \textbf{0.495} &   \textbf{0.658} &        \textbf{0.639} \\
  \midrule
 \multirow{3}{*}{\atomic{}} & \gptxl{}     &    0.083 &    0.029 &    0.011 &    0.005 &    0.081 &     0.087 &   0.045 &        0.386 \\
 & \cometgptxl{}        &    0.419 &    0.296 &    \textbf{0.228} &    \textbf{0.189} &    0.292 &     0.517 &   0.733 &        0.634 \\
 & \cometbart{}            &    \textbf{0.515} &    \textbf{0.324} &    0.220 &    0.159 &    \textbf{0.347} &     \textbf{0.546} &   \textbf{0.740} &        \textbf{0.646} \\
  \midrule
 \multirow{3}{*}{\conceptnet{}} & \gptxl{} &    0.044 &    0.012 &    0.004 &    0.002 &    0.064 &     0.057 &   0.050 &        0.389 \\
 & \cometgptxl{}     &    0.155 &    \textbf{0.119} &    \textbf{0.095} &    \textbf{0.078} &    \textbf{0.134} &     \textbf{0.193} &   \textbf{0.425} &        \textbf{0.552} \\
 & \cometbart{}        &    \textbf{0.172} &    0.111 &    0.072 &    0.049 &    0.130 &     0.184 &   0.368 &        0.535 \\
  \midrule
 \multirow{3}{*}{\transomcs{}} & \gptxl{} &    0.028 &    0.001 &    0.000 &    0.000 &    0.093 &     0.053 &   0.013 &        0.351 \\
 & \cometgptxl{}     &    0.301 &    0.000 &    0.000 &    0.000 &    0.180 &     0.302 &   0.254 &        0.677 \\
 & \cometbart{}         &    \textbf{0.351} &    \textbf{0.170} &    \textbf{0.003} &    0.000 &    \textbf{0.198} &     \textbf{0.352} &   \textbf{0.297} &        \textbf{0.678} \\
 \bottomrule
\end{tabular}
\caption{Automated metrics for the quality of the \emph{tail} generations of the \gptxl{} language model and the knowledge models \cometgptxl{} and \cometbart{}. Each approach uses greedy decoding for  sampled 5k test prefixes for each KG. The 5k prefixes correspond to the ones for the human eval. Similar results are obtained on the full test sets (cf. Appendix~\ref{app:neural}).}

\label{tab:auto-eval}
\end{table*}

\vspace*{2mm}
\noindent
\textbf{Experimental Setup.} To evaluate whether knowledge graphs can help language models effectively transfer to \textit{knowledge models}, we train different pretrained language models on the knowledge graphs described in Section~\ref{sec:kb-comparison-coverage}, which we describe below:

\noindent \textbf{GPT2} \cite{radford2019language} is a Transformer \cite{Vaswani2017AttentionIA} based language model. In our experiments, we use the largest GPT2 model, \gptxl{}, that has 1.5B parameters. We fine-tune \gptxl{} on each of our CSKGs to predict the tail of a tuple (e.g., wheat) given the head (e.g., bread) and a relation (e.g., \MadeUpOf). The hyperparameter settings used for training are described in more detail in Appendix.
Additionally, we use \gptxl{} in a zero-shot setting as a baseline to measure the effect of transfer learning on knowledge graphs. For fair comparison, we convert each relation manually to an English language prompt expecting the tail of each tuple as output generated by the model.

\noindent \textbf{BART} \citep{lewis-etal-2020-bart} is a Bidirectional and Autoregressive Transformer, 
an adaptation from BERT \cite{bert} that is better suited 
for natural language generation (e.g., translation, summarization). 
Additional training details are provided in Appendix~\ref{app:neural}.

\noindent \textbf{GPT-3} \citep{brown2020language} is an autoregressive language model that has 175B (over 100X more parameters than \gptxl{}) parameters and is trained on a corpus of web text. We use the \gpttt{} API to \textit{prime} the language model to generate the tail for a given prefix -- \textit{(head, relation)} pair. Thus, \gpttt{} is evaluated in a few-shot setting. Additional details of our implementation are provided in Appendix~\ref{app:neural}.

\vspace*{2mm}
\noindent
\textbf{Evaluation Setup.}
To assess language-to-knowledge transfer capabilities, we evaluate how language models generalize to new, unseen entities, concepts, or events. We split each knowledge graph into training, validation, and test sets such that the \emph{heads} of the knowledge tuples do not overlap between these sets. This adversarial split forces the language models to generalize the relationships they learn from training on the knowledge graphs to the entities learned during language pretraining. 
Also, to avoid overpopulating the validation and test sets with generic \emph{heads} (e.g., ``I'', ``You'', ``He'', ``We'', and ``They'' collectively account for over 2.2M tuple heads in \transomcs), we enforce that the head of any knowledge tuple in the \emph{dev} and \emph{test} sets is involved in at most $500$ tuples. Finally, we remove low-quality tuples from \transomcs{} by imposing a confidence score of $\geq 0.5$. 

We score the tuples generated by these knowledge models using common evaluation metrics for text generation: BLEU \citep{papineni2002bleu}, ROUGE \citep{lin2004rouge}, CIDEr \citep{vedantam2015cider}, and BERT Score \citep{zhang2019bertscore}. For a subset of 5000 generated tuples from the test set of each knowledge graph, we also run the same human evaluation described in Section~\ref{sec:kb-comparison-precision}.

\vspace*{2mm}
\noindent
\textbf{Results.} We present our main results in Tables~\ref{tab:human-eval-generations} and~\ref{tab:auto-eval}. First, we note the large divide between the zero-shot \gptxl{} model that produces commonsense knowledge without any fine-tuning and the two \comet models across the \atomicTT, \atomic, and \conceptnet knowledge graphs (Table~\ref{tab:human-eval-generations}). This large gap indicates that language models can benefit from learning facts from commonsense knowledge graphs. They do not have the means to precisely express this knowledge directly from just pretraining on language. This observation is supported by the gaps between these models in the automatic evaluations (Table~\ref{tab:auto-eval}), as well. Additionally, human evaluation of \gpttt{} (Table~\ref{tab:human-eval-generations}) shows a $\sim$12 point deficit compared to the performance of \cometbart{}, in spite of \gpttt{} (175B) having over $\sim$430 times more parameters than \cometbart{} (406M). Similarly, we see a large gap in performance across all automated metrics in Table \ref{tab:auto-eval}. The performance gap indicates that high-quality declarative knowledge is valuable even after the advent of extreme scale language models. 

In addition to this main result, two particularly interesting observations emerge. First, we note that the gap between the zero-shot model and \comet is larger on the \atomicTT{} 
and \atomic{} knowledge graphs, than on \conceptnet 
, supporting the reflection that \atomicTT{} supports categories of knowledge that are more difficult to learn from pretraining. 
Second, the results on the human evaluation show that \comet models trained on \transomcs{} are not able to generalize knowledge to new entities, implying that language models benefit more from accurate knowledge examples, which \transomcs lacks (cf. \S\ref{sec:kb-comparison-precision}).

%% file: sections/acknowledgements.tex
We would like to thank the anonymous reviewers for their valuable feedback.
This research was supported in part by NSF (IIS-1524371), the National Science Foundation Graduate Research Fellowship under Grant No. DGE 1256082, DARPA CwC through ARO (W911NF15-1- 0543), DARPA MCS program through NIWC Pacific (N66001-19-2-4031), and the Allen Institute for AI. 
Computations on \url{beaker.org} were supported in part by credits from Google Cloud. TPU machines for conducting experiments
were provided by Google.

%% file: sections/appendix.tex

\section{\atomic2020{} Details}
\label{appendix:atomic2020-details}

\subsection{\atomicTT{} Relations}

\begin{figure*}[t]
\centering
\includegraphics[width=.65\textwidth]{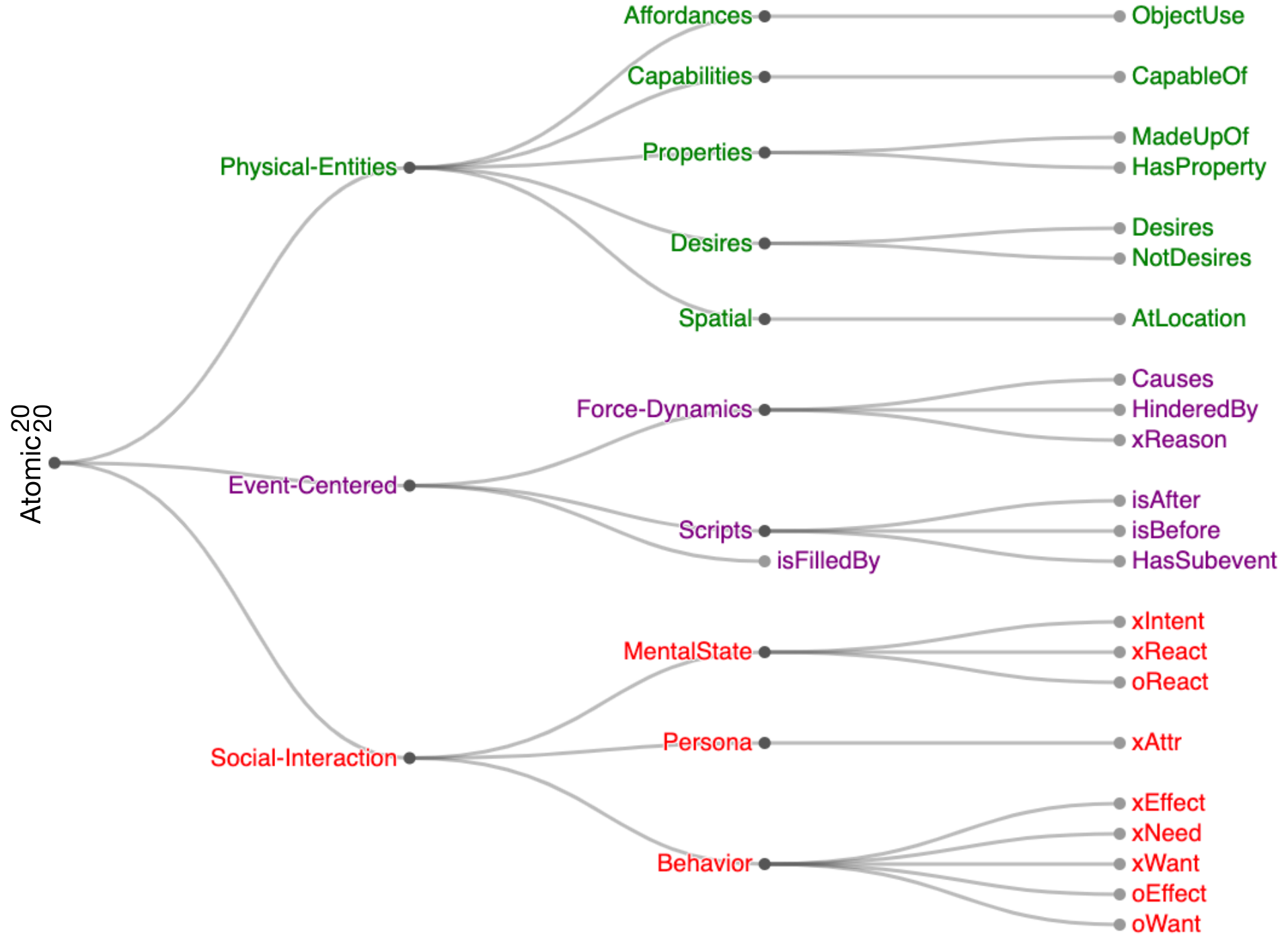}
\caption{\atomicTT{} relations organized into a hierarchical structure.}
\label{fig:hierarchy}
\end{figure*}

In this section we detail the relations in \atomicTT{}. Figure~\ref{fig:hierarchy} shows the hierarchical breakdown of the \atomicTT{} relation labels. While there is no internal structure directly encoded for \atomicTT{} relations, they fall into three natural categories based on their meaning:  physical-entity, social-interaction  and event-centered commonsense.


\vspace{2mm} \noindent \textbf{Physical-Entity Commonsense.}
Physical-entity commonsense deals with inferential knowledge about common entities and objects. Physical commonsense such as these is crucial for interacting with the world: allowing us to distinguish the dangerous (e.g., ``fire can be painful'') from the harmless (e.g., ``teddy bears are comforting''), manipulate objects for our use (e.g., ``helmets protect head''), and solve problems (e.g., ``how do I open this door?''). We identify \textbf{seven} relations under this category.

\begin{figure*}
\centering
\fbox{\includegraphics[width=.95\linewidth]{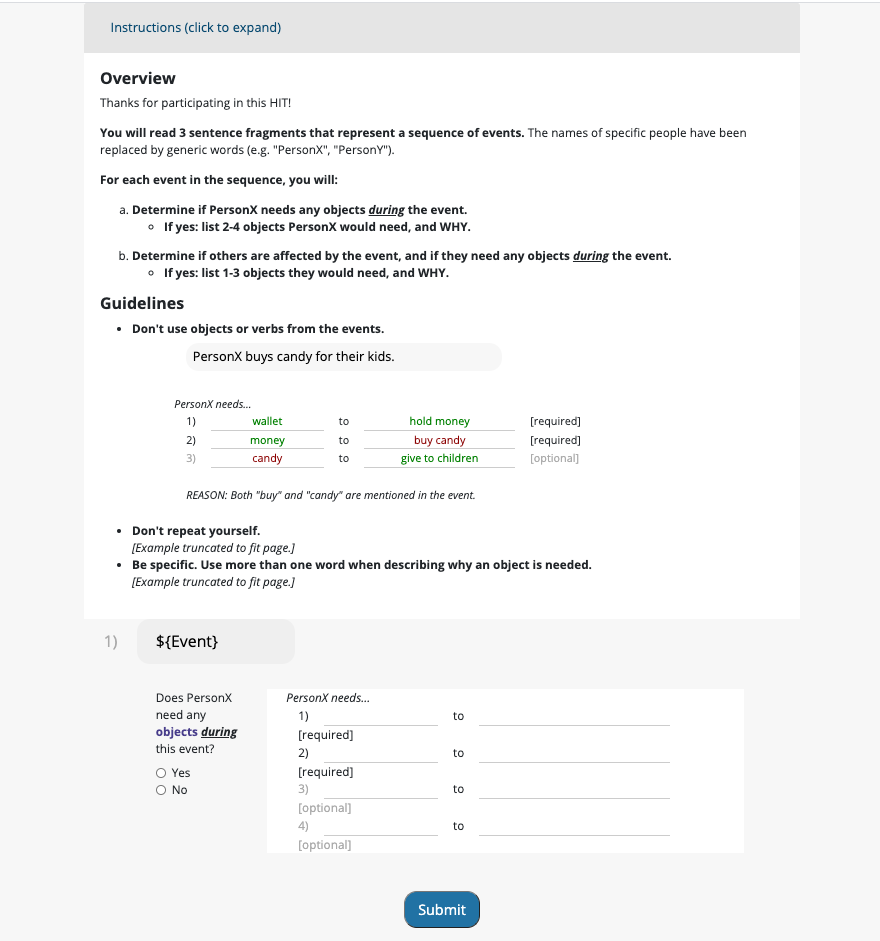}}
\caption{Mechanical Turk template used to collect ObjectUse tuples. We collected three sets of object affordances per HIT, but two have been truncated to fit this page.}
\label{fig:objectuse_template}
\end{figure*}

\begin{itemize}
\item \textbf{\ObjectUse{}} describes everyday \textbf{affordances} or uses of objects, and includes both \textit{typical} and \textit{atypical} uses. For example, ``popcorn bucket'' can typically be used to ``hold popcorn'' but it could also serve ``as a hat'' in atypical situations. The template used to collect object affordances is shown in Figure \ref{fig:objectuse_template}.

\item \textbf{\MadeUpOf{}} and \textbf{\HasProperty{}}, two \textbf{property} relations,  denote the relationship between an entity and its composition or characteristics. \MadeUpOf{} describes a part, portion or makeup of an entity. For example, ``cake'' can be \MadeUpOf{} ``eggs'' (composition/ingredient) or ``icing'' (part/portion). Similarly, \HasProperty{} usually describes entities' general characteristics such as ``rose'' is ``red,'' subjective attributes such as ``thirst'' is ``uncomfortable.'' In certain case, the relation can also map to descriptors that speak to the substance or value of items such as ``meat'' has property of being ``stored in the freezer'' or ``bike'' is ``powered by person's legs.''

\item \textbf{\AtLocation{}} is a \textbf{spatial} relation that describes the location in/on/at which an entity is likely to be found (e.g. ``gambler'' can be found in ``casino,'' ``wrench'' can be found in ``garage'').

\item \textbf{\CapableOf{}} is designed to describe abilities and \textbf{capabilities} of everyday living entities (e.g., humans, animals, insects) and natural entities that can exert a force (e.g. sun, storms). \CapableOf{} includes general capabilities such as a ``human'' is capable of ``thinking and reasoning'' or ``drinking coffee.'' It also includes specialized capabilities such as a ``surgeon'' is capable of ``operating on a patient.'' 

\item \textbf{\Desires{}} and \textbf{\NotDesires{}} are relations that deal with \textbf{desires}\footnote{Since desire relations are about cognitive states of sentient beings, they also provide a degree of commonsense about social-interaction. However, we point out that these relations indicate \textit{generic} characterizations of animate entities rather than describing situationally-based cognitive mental states (e.g., X being `encouraged' only applies to the event it is situated in). For this reason, we include these relations under physical-entity commonsense.
} 
of sentient entities; e.g., ``doctors'' likely desire to ``cure patient'' but do not desire ``malpractice suit.'' 

\end{itemize}

\vspace{2mm} \noindent \textbf{Social-Interaction Commonsense.}
Social-interaction relations comment on socially-triggered states and behaviors. Social commonsense is useful for gauging people's intentions and purpose, and predicting situationally-relevant human reactions and behaviors. Following the definitions for \atomic{} relations \cite{sap2018atomic}, we identify a total of \textbf{nine} relations within this category. 

\begin{itemize}
\item Three \textbf{mental state} relations address the emotional or cognitive states of the participants in a given event. \textbf{\xIntent{}} defines the likely \textit{intent} or desire of an agent (X) behind the execution of an event. Given the head ``X gives Y gifts,'' an \xIntent{} might be that X wanted ``to be thoughtful.'' Relations \textbf{\xReact{}} and  \textbf{\oReact{}} define the \textit{emotional reactions} on the part of X or other participants in an event. As a result of gift giving, X might feel ``good about [one]self'' and others (in this case, Y) might feel ``appreciated.''

\item Five \textbf{behavioral} relations address the socially relevant responses to an event. \textbf{\xNeed{}} describes a precondition for X achieving the event. For example, in order for X to give Y gifts, X must first ``buy the presents.'' \textbf{\xWant{}} and \textbf{\oWant{}} are postcondition desires on the part of X and others, respectively. As a result of X giving Y gifts, X may also desire ``to hug [Y]'' and Y may want to ``open the gift.'' \textbf{\xEffect{}} and \textbf{\oEffect{}} are social actions that may occur after the event: X may ``get hugged'' and Y may ``blush'' in response.

\item The last relation \textbf{\xAttr{}} describes X's \textbf{persona} or attribute as perceived by others given an event. In the gift giving example, X may be seen as ``generous'' or ``giving.'' In contrast, in an event such as ``X steals a car,'' X may be perceived as ``evil.''  

\end{itemize}

\vspace{2mm} \noindent \textbf{Event-Centered Commonsense.}
While social-interaction commonsense gauges human behaviors and mental states given an event, the event-centered commonsense provides intuitions about how common events are related to one another. Commonsense about event interaction is useful for understanding likely causes and effects of events in the world. This knowledge allows humans to strategize and explore the best solutions for their objectives, make contingency plans, and revise goals when circumstances deviate from expectation. There are \textbf{seven} relations that fall under this category.

\begin{figure*}
\centering
\fbox{\includegraphics[width=.95\linewidth]{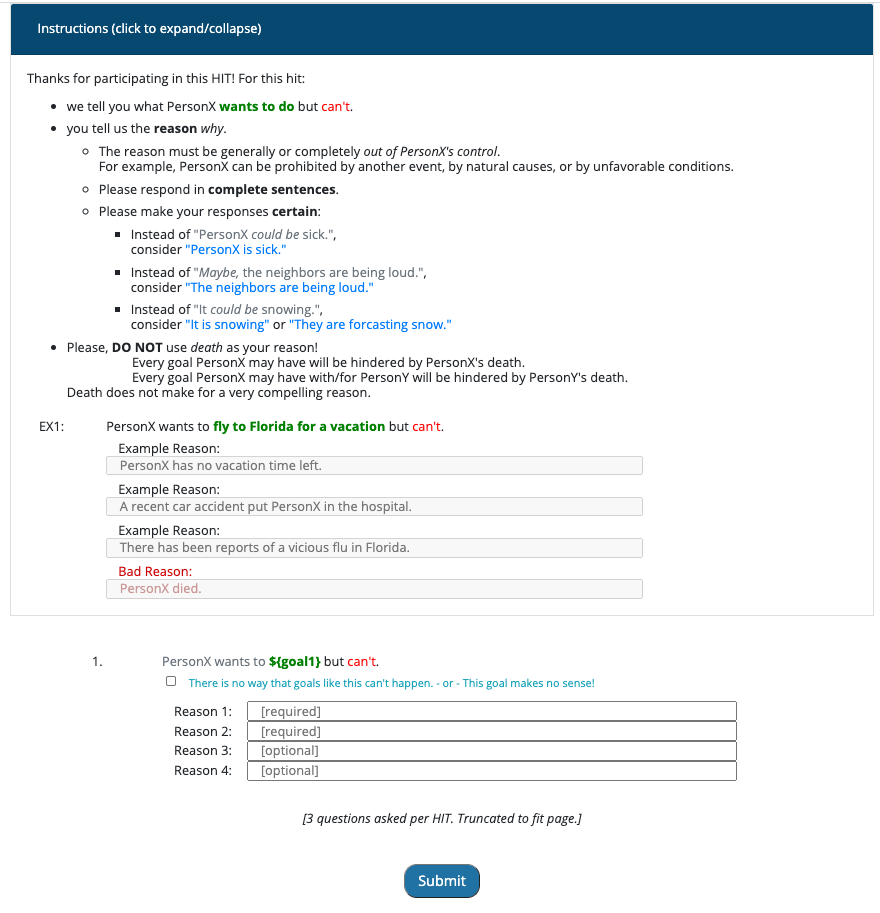}}
\caption{Mechanical Turk template used to collect HinderedBy tuples. We collected three sets for hindrances of goals per HIT, but two have been truncated to fit this page.}
\label{fig:hinderedby_template}
\end{figure*}

\begin{itemize}
\item We group three relations under \textbf{force dynamics}.\footnote{For a discussion of force dynamics in cognitive linguistic and lexical semantic literature cf. \citet{Herskovits2009LanguageAS,landau1991spatial,Talmy1988ForceDI}.} This group conceptualizes dynamic interactions between events with regards to exerted causal forces and impelled actions. \textbf{\Causes{}} specifically captures the causal relation between two events or entities -- e.g. an ``accident'' can cause ``injury.'' \Causes{} does have some overlap with behavioral relations such as \xEffect{} in that they are postconditions of an event, but the postcondition in \Causes{} is not socially triggered and can exist outside human control (e.g., ``bad weather''  causes ``power outages''). 
\textbf{\HinderedBy{}} introduces hindrances that obstruct the natural path to the achievement of a goal. For example, the event ``X adopts a cat'' can be obstructed if ``X is allergic to cats.''  \textbf{\xReason{}} provides a post-fact explanation of the cause of an event (e.g., why one has to ``walk'' could be explained by ``car has broken down''), which is related to, but distinct from, \xIntent{}'s intentions (i.e., ``X walks'' because X wanted to ``go home''). The template used to collect goal hindrances is shown in Figure \ref{fig:hinderedby_template}.

\item Three relations provide reasoning about event \textbf{scripts} or sequences.  \textbf{\IsAfter{}} and \textbf{\IsBefore{}} introduce events that can precede or follow an event, respectively. For example, ``X is in a hurry to get to work'' can happen after ``X wakes up 15 minutes late'' and before ``X drives too fast.'' These relations are distinguished from behavioral relations \xNeed{} (pre-condition) and \xEffect{} (post-condition) in that \IsAfter{} and \IsBefore{} are temporally situated without specific regard to the need or reaction of the person X. For example, ``X pushes X's luck'' can happen before ``X gets a broken nose'' but getting a broken nose is not an action X intentionally may take after pushing one's luck.
Relation \textbf{\HasSubEvent{}} provides the internal structure of an event, each tail denoting a step within the larger head event.

\item The last relation in the event-centered category, \textbf{\isFilledBy{}}, provides a filler phrase for an event with a blank that is sensical and commonly acceptable for the event. For example, the blank in an event such as ``X catches \_\_\_ in the act'' can be commonly filled by entities such as a ``cheater,'' a ``burglar,'' or a ``mouse.''



\end{itemize}

\subsection{\atomicTT{} Tuples}
In this section, we detail the population of the \atomicTT{} tuples (see Table \ref{tab:atomic2020-newrelations} for counts per relation).

\vspace{2mm} \noindent \textbf{Social-Interaction Tuples.}
For social-interaction relations, we incorporated $877$K tuples from \atomic{}, and crowdsourced an additional $34$K tuples using the same approach as \citet{sap2018atomic}. The rest of this section will refer to the head events in the social-interaction tuples as \textbf{base events}.

\vspace{2mm} \noindent \textbf{Crowdsourced Tuples.} Tuples for relations \ObjectUse{}, \HinderedBy{}, \isFilledBy{}, \IsBefore{} and \IsAfter{} were crowdsourced via Amazon Mechanical Turk.  We paid an average of \$$15$ an hour for our crowdsourcing efforts. We release all crowdsourcing templates as part of our codebase.\footnote{\url{http://anonymous}}

\begin{itemize}
\item For the collection of \HinderedBy{}, we crowdsourced over 100K event-hindrance tuples by prompting the workers with base events from \atomic{} and eliciting reasons why one may not be able to achieve the event. In order to make the prompt events more worker-friendly, we processed the events as a desire (e.g., ``X adopts a cat'' $\rightarrow$ ``X \textit{wants to} adopt a cat'').\footnote{To achieve this, we removed modal verbs, lemmatized the head verb of the sentence, and inserted a `want to' phrase before the verb.} We specifically elicited personal causes (e.g., ``X is allergic to cats''), situational causes (e.g., ``there are no pet stores nearby''), and social causes (e.g., ``X's landlord disallows pets''). 

\item $33$K \isFilledBy{} tuples were collected by presenting workers with base events. The workers were asked to provide two (up to four) common objects or entities that will make sense in the sentence.  

\item $46$K tuples for \IsBefore{} and \IsAfter{} were collected together as sequences of events. Given a base event, the workers were asked to write a short 3-sentence story by providing a preceding and following event. The workers were given the option to opt out of writing a story if they felt that the event they were given didn't make sense enough to create a story. 

\item As discussed in the main text (\S\ref{sec:atomicTT}), 
$130$K  \ObjectUse{} tuples were crowdsourced by eliciting common objects and their uses for every event in the collected event sequences. For each event, a worker was asked to provide 2-4 common items that were needed during the displayed event. Atypical \ObjectUse{} was collected in a second pass, where for each collected unique object, the workers were prompted with the object and asked to provide an atypical, creative or resourceful use for the item shown.
\end{itemize}

\begin{table}[t]
\small
\center
\begin{tabular}{ll}
\toprule
 \textbf{\conceptnet{}} &      \textbf{\atomicTT{}}                       \\
\midrule
AtLocation      & AtLocation \\
CapableOf      & CapableOf \\
Causes            & \textbf{Causes}, xEffect                         \\
CausesDesire     & xWant   \\
Desires         & Desires                                        \\
MadeOf          & MadeUpOf \\
HasProperty    & HasProperty         \\
HasA            & \textbf{MadeUpOf}, HasProperty     \\
HasPrerequisite & xNeed \\
HasSubevent     & HasSubEvent \\
HasFirstSubevent & HasSubEvent \\
HasLastSubevent & HasSubEvent \\
NotDesires      & NotDesires                                     \\
MotivatedByGoal & \textbf{xIntent}, xReason                                \\
PartOf          & MadeUpOf \\
UsedFor         & ObjectUse                        \\
ReceivesAction  & \makecell[lt]{\textbf{MadeUpOf}, AtLocation, \\ Causes, ObjectUse}  \\
\bottomrule                                  
\end{tabular}
\caption{\conceptnet{} relations mapped to \atomicTT{} relations. For labels mapping to multiple \atomicTT{} relations, the one that received the majority mapping is bolded. }
\label{tab:conceptnet-atomic-relation-mapping}
\end{table}

\begin{table}[t]
\small
\begin{tabular}{ll}
\toprule
\textbf{Relations}       & \textbf{Human Readable Template}         \\
\midrule
AtLocation               & located or found at/in/on                \\
CapableOf                & is/are capable of                        \\
Causes                   & causes                                   \\
CausesDesire             & makes someone want                       \\
CreatedBy                & is created by                            \\
Desires                  & desires                                  \\
HasA                     & has, possesses or contains               \\
HasFirstSubevent         & BEGINS with the event/action             \\
HasLastSubevent          & ENDS with the event/action               \\
HasPrerequisite          & to do this, one requires                 \\
HasProperty              & can be characterized by being/having     \\
HasSubEvent              & includes the event/action                \\
HinderedBy               & can be hindered by                       \\
InstanceOf               & is an example/instance of                \\
isAfter                  & happens after                            \\
isBefore                 & happens before                           \\
isFilledBy               & blank can be filled by                   \\
MadeOf                   & is made of                               \\
MadeUpOf                 & made (up) of                             \\
MotivatedByGoal          & is a step towards accomplishing the goal \\
NotDesires               & do(es) NOT desire                        \\
ObjectUse, UsedFor       & used for                                 \\
oEffect                  & as a result, Y or others will            \\
oReact                   & as a result, Y or others feels           \\
oWant                    & as a result, Y or others want            \\
PartOf                   & is a part of                             \\
ReceivesAction           & can receive or be affected by the action \\
xAttr                    & X is seen as                             \\
xEffect                  & as a result, PersonX will                \\
xIntent                  & because PersonX wanted                   \\
xNeed                    & but before, PersonX needed               \\
xReact                   & as a result, PersonX feels               \\
xReason                  & because                                  \\
xWant                    & as a result, PersonX wants               \\
\bottomrule
\end{tabular}
\caption{Human readable templates for each relation used for crowdsourced human evaluations.}
\label{tab:human-readable-templates}
\end{table}

\vspace{2mm} \noindent \textbf{Integration of \conceptnet{} Tuples.} The tuples for the remaining relations are populated through the integration of the commonsense portion of \conceptnet{}. As discussed in the main text, a select subset of \conceptnet{}(v5.7) tuples ($172$K) were integrated into \atomicTT{}. 

The primary challenge in integrating \conceptnet{} tuples into \atomicTT{} was in identifying knowledge that is most likely to reflect commonsense information. \conceptnet{}(v5.7) contains tuples built on not only concept relationships directly sourced from human informants, but also on information pulled from other lexical sources such as WordNet \cite{Miller1995} and DBpedia \cite{Auer2007DBpediaAN}, which automatically extracts knowledge from Wikipedia articles \cite{speer2017conceptnet}. As a result, even those relations that are \textit{designed} to primarily represent commonsense knowledge (i.e., the OMCS relations) include among the mix, tuples that reflect factual or lexical co-occurrence knowledge. These examples deviate from the type of knowledge we would ideally consider as ``commonsense,'' i.e., qualitative experiential knowledge gained through subjective observation of and interaction with the world. Relations such as \InstanceOf (``is instance/example of'') stands as a case in point (e.g., ``tortilla'' is an example of ``flatbread'' or ``toffee'' is an example of ``candy''). While included within the OMCS relations, the encoded information can be hard to distinguish from the more accepted taxonomic relations such as \IsA{} (``is a kind/type of'').\footnote{In fact, \conceptnet{}(v5.7) recognizes the similarities between \IsA{} and \InstanceOf{} and has accordingly deprecated \InstanceOf{} in favor of \IsA{}. Nevertheless, \InstanceOf{} is still found in \conceptnet{}(v5.7).} Relationships found in relations such as \RelatedTo{} and \DistinctFrom{} are too underspecified with regards to the meaning they represent, and for other relations such as \LocatedNear{}, and negative forms such as \NotCapableOf{} or \NotHasProperty{}, the relationships amount to general lexical relationships.

Thus, the process of \conceptnet{}(v5.7) knowledge selection (described in \S\ref{sec:atomicTT}) was judiciously guided by three competing priorities: when possible, we prioritized (1) qualitative commonsense over factual knowledge, (2) general knowledge over highly specific knowledge (e.g., personal names), and (3) meanings that are specific enough to be meaningfully categorized. Since the ideal route of verifying imported data via crowdsourcing can be resource-intensive, we opted for an approach whereby relations were first selected based on the data they represent; then tuples were pruned based on heuristics that leverage lexical and syntactic information of the concepts. As mentioned in the main text, $10$\% of the data selected for integration was validated by crowdworkers, yielding a greater than $93$\% acceptance rate. Three relations, namely \HasProperty{}, \CapableOf{}, and \MotivatedByGoal{}, were sent for instance-by-instance crowdsourcing for the purpose of debiasing human-related descriptions, and subdividing semantically distinct elements within the category (e.g., \MotivatedByGoal{} mapped to \xIntent{} and \xReason{}). The resulting \conceptnet{}-to-\atomicTT{} relation mapping details are shown in Table~\ref{tab:conceptnet-atomic-relation-mapping}.

\begin{figure*}[ht]
\centering
\includegraphics[width=0.35\linewidth]{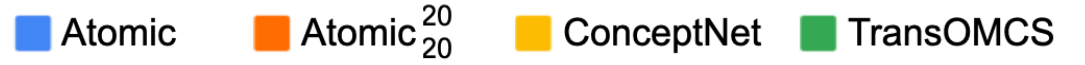}\\
\includegraphics[trim=20 20 20 20,clip,width=0.33\linewidth]{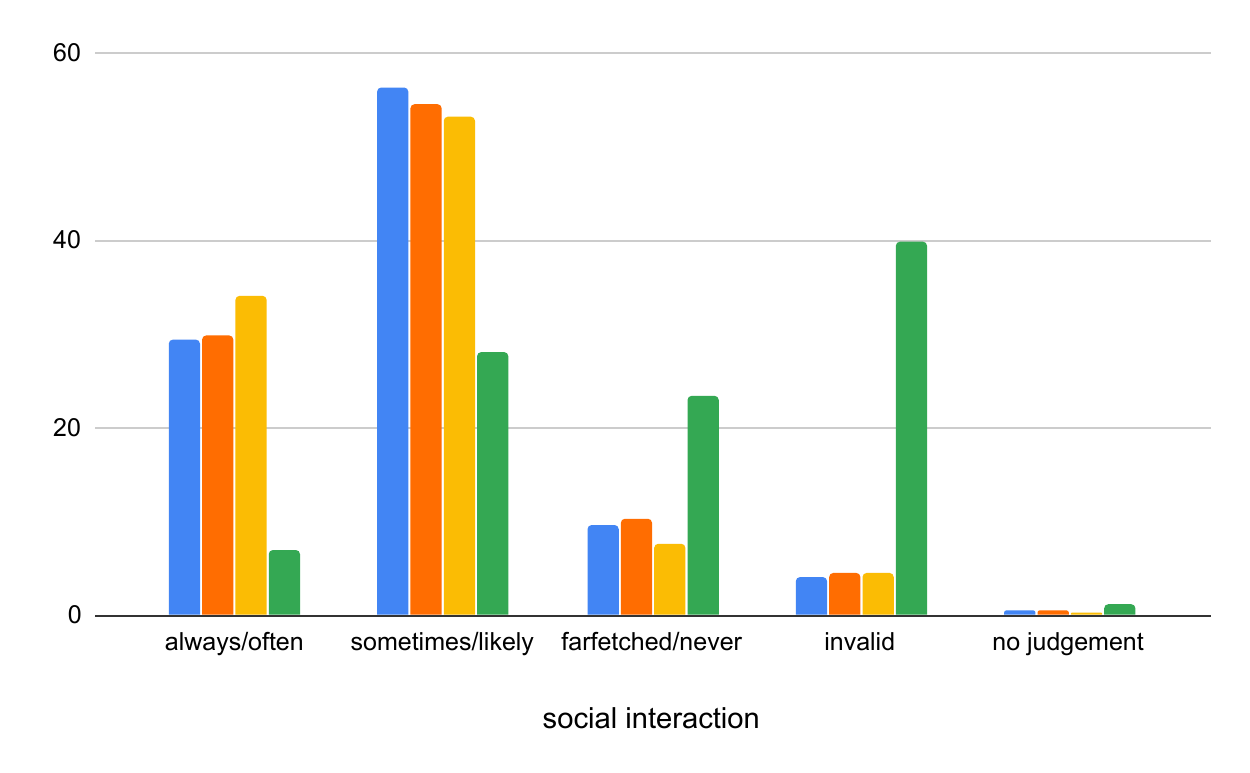}
\includegraphics[trim=20 20 20 20,clip,width=0.33\linewidth]{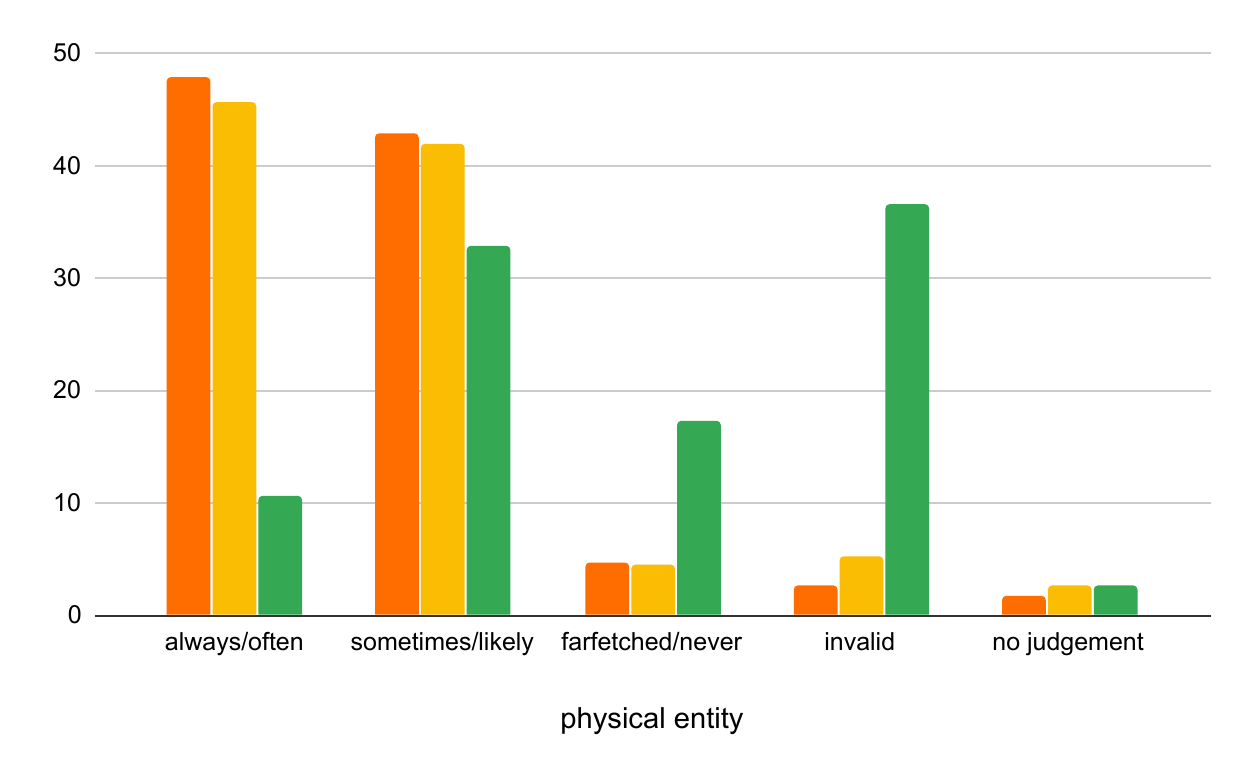}
\includegraphics[trim=20 20 20 20,clip,width=0.33\linewidth]{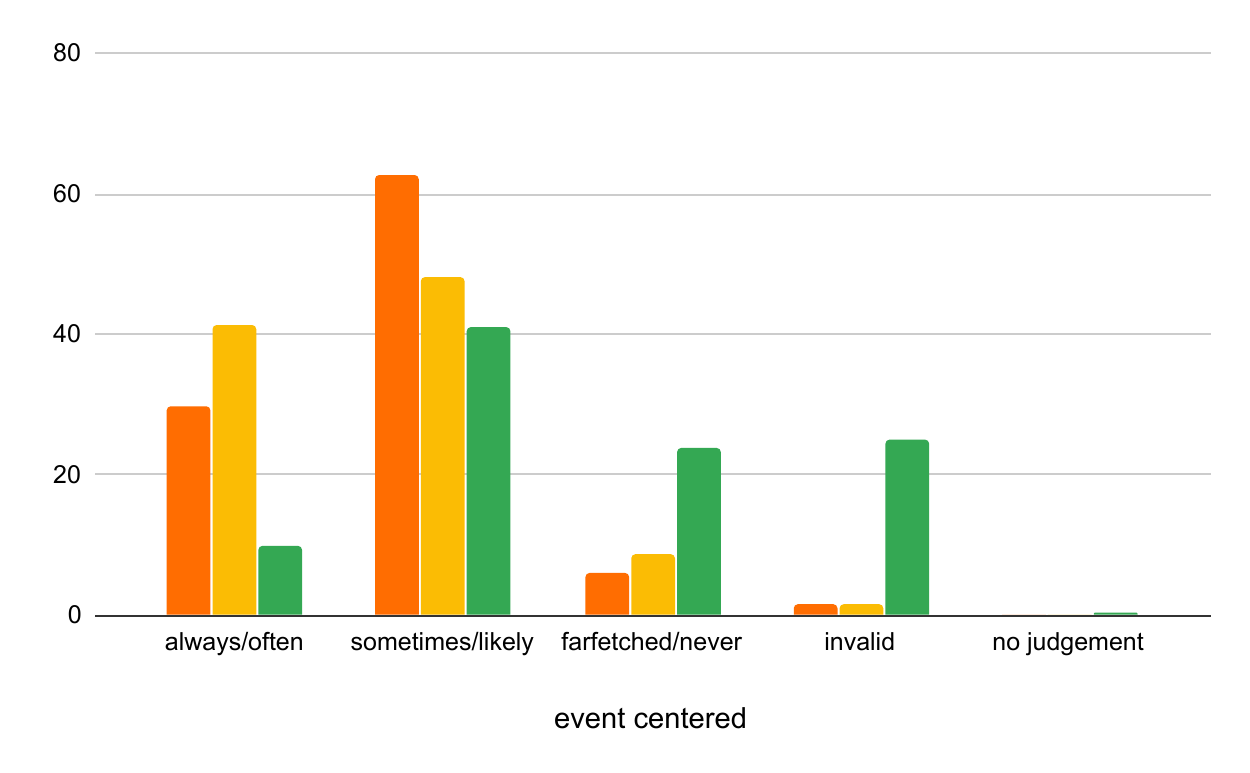}
\caption{Percentage distribution of raw accuracy ratings broken down by KB (i.e., breakdown of Table~\ref{tab:precision-results}). From left to right are the ratings for social-interaction tuples, physical-entity tuples, and event-centered tuples. We use the \conceptnet{}-to-\atomicTT{} relation mappings (shown in Table~\ref{tab:conceptnet-atomic-relation-mapping}) to categorize \conceptnet{} and \transomcs{} relations into the three categories. For multiple mappings, we map the \conceptnet{}/\transomcs{} labels to the majority mapped label (in bold in Table~\ref{tab:conceptnet-atomic-relation-mapping}). Note that the latter two figures do not include \atomic{} as the KB only includes social-interaction relations.}
\label{fig:accuracy-breakdown}
\end{figure*}

\section{Symbolic Knowledge Graph Details}
\label{app:evaluation-details}

\subsection{Human Evaluation}
 
\noindent \textbf{Human Readable Relation Templates.} Since the KB relation labels are rather telegraphic on their own, we used human readable language forms (based \atomicTT{} and \conceptnet{} definitions) for prompt display in crowdsourced evaluations. The complete list is available in Table~\ref{tab:human-readable-templates}.

\subsection{KB Accuracy \& Coverage}
\noindent \textbf{In Table~\ref{tab:precision-results}, what type of tuples generally end up with no judgment?} Tuples receiving \textit{no judgment} fall into three general categories: (1) either the head or the tail concept is too specialized for the workers to judge without consulting reference (e.g., ``klebsiella'' is part of ``bacteria,'' ``drug cocktail'' made of ``nucleoside reverse transcriptase inhibitor''); (2) concepts refer to highly specific entities or referents (e.g., ``singh'' capable of ``bring key,'' ``falkland island islas malvinas'' part of ``argentina''); and (3) \textit{Reject} candidates that workers have decided to hedge on (e.g., ``dandelion'' used for ``love,'' ``democrat'' desires ``matter''). Such tuples are mostly found in TransOMCS, as evidenced by the high fraction of tuples that received \textit{No Judgment} at less than half of \atomicTT's \textit{Accept} rate (see \ref{tab:precision-results}).

\vspace{2mm} \noindent \textbf{Accuracy reported in Table~\ref{tab:precision-results} for \transomcs{} is based on the complete set. What do the numbers look like for the top 1\% and top 10\% of \transomcs{}?} The evaluation for top 1\% and top 10\% are indeed higher than the reported values for \transomcs{}. However, they still lag behind other KBs (Table \ref{tab:precision-results-omcs}).

\begin{table}[t]
\center
\small
\begin{tabular}{lrrrr}
\toprule
\textbf{$\transomcs$} & \textbf{Accept}	& \textbf{Reject} & \makecell{\textbf{No}\\ \textbf{Judgment}} & \makecell{\textbf{Data}\\ \textbf{ Fraction}} \\
\midrule
Top 1  & 65.3 & 31.1 & 3.6 & 7.2\% \\
Top 10  & 49.1 & 45.7 & 5.1 & 26.1\% \\
All (reported)  & 41.7 & 53.4 & 4.9 & 100\%\\
\bottomrule
\end{tabular}
\caption{\transomcs{} \textbf{accuracy} for top Top 1\% and Top 10\% subsets for the corresponding results reported in Table~\ref{tab:precision-results}. }
\label{tab:precision-results-omcs}
\end{table}

\vspace{2mm} \noindent \textbf{Does the accuracy ratings breakdown for each KB provide further insights?} A closer look at the raw accuracy ratings shows an interesting emergent rating pattern across KBs (Table~\ref{tab:precision-results}). For all KBs with the exception of \transomcs{}, we observe that the majority of social-interaction \textit{Accept} originate from the  \textit{sometimes/likely} rating. However, such preference is not seen in the physical-entity tuples, which show a slightly higher tendency for the \textit{always/often} rating. For event-centered tuples, \atomicTT{} favors the \textit{sometimes/likely}, while \conceptnet{} does not. \transomcs{} shows highest ratings for the \textit{sometimes/likely} and \textit{invalid} ratings, and the patterns are invariant across the board.

One additional point to mention is that \atomicTT{} social-interaction and event-centered tuples proportionally contain more of the human-crowdsourced commonsense knowledge than the physical-entity category, which, with the sole exception of \ObjectUse{}, includes tuples integrated from \conceptnet{} graph. 
The observation that much of the knowledge in \atomicTT{} is sometimes or likely true, reflects our intentional efforts to deprioritize factual information over qualitative commonsense knowledge. More importantly, it shows that most of the knowledge within the \atomicTT{} graph can be, under the right circumstances, defeasible. That is, one can pose a likely hypothesis that a hindrance to ``X writes stories'' is that ``X can't read;'' however, such a hypothesis can be defeated if we also know that X has written stories before. We find that such context-dependent ambiguities are of more compelling interest to us, as certainties may be better covered by language models.



\section{Neural Knowledge Graph Details}
\label{app:neural}

\vspace*{2mm}
\noindent
\textbf{Dataset Split.}
Table \ref{tab:split_sizes} reports the number of tuples for each three-way split (train/dev/test) of each knowledge graph. The \atomicTT{} split preserves the splits from \atomic and \conceptnet: any tuple in \atomicTT that appears in the train (resp. dev, test) set of \atomic or \conceptnet belongs to the train (resp. dev, test) set of \atomicTT. Overall, \atomicTT{} provides over $50\%$ more tuples than the initial version \atomic.

\input{tables/tab-split-sizes}

\input{tables/tab-automated-metrics-full-test}

\begin{figure*}
\centering
\includegraphics[width=.95\linewidth]{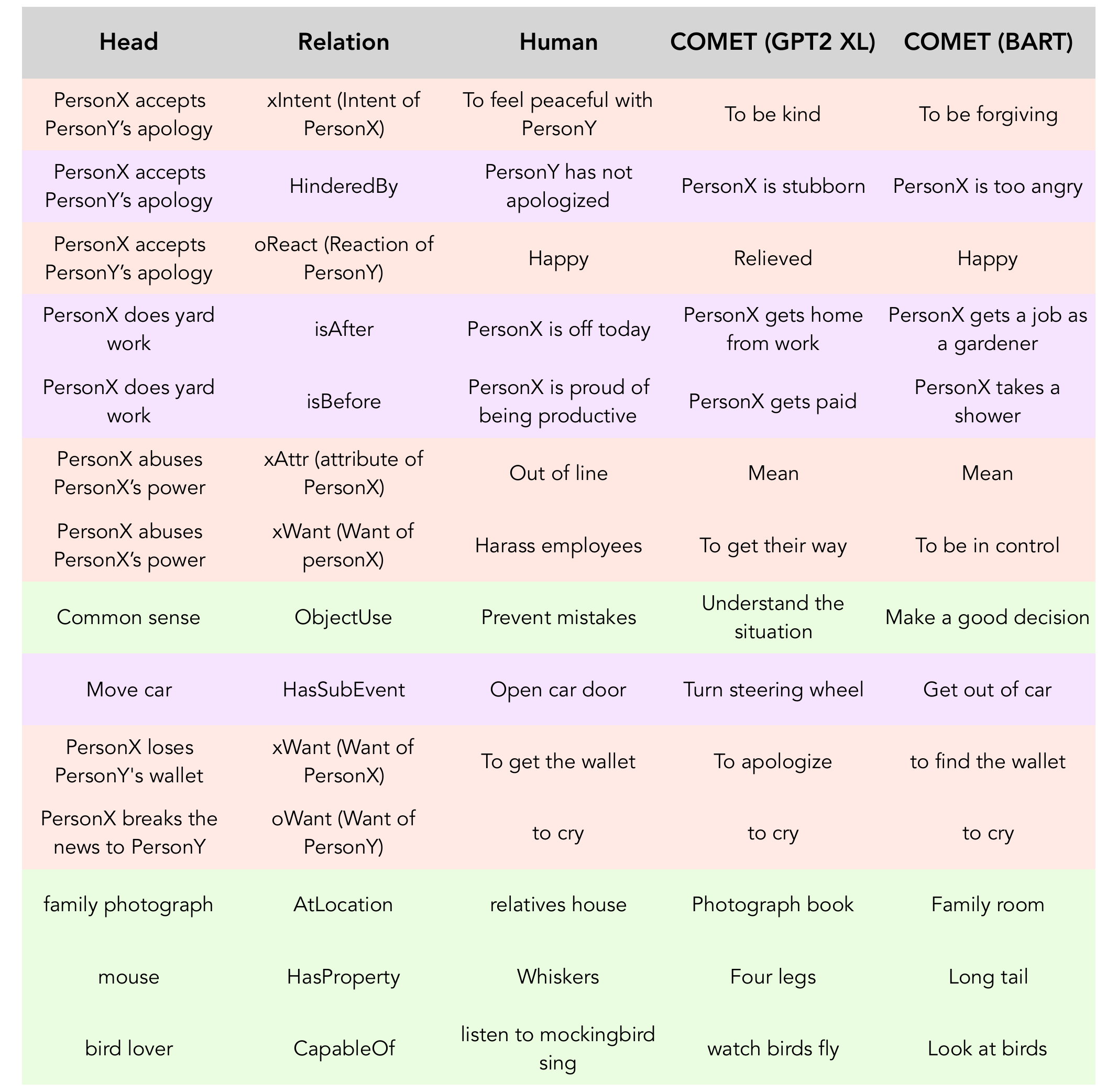}
\caption{Example generations of models on relations from \atomicTT{}. \textcolor{myred}{Red}, \textcolor{violet}{purple} and \textcolor{darkgreen}{green} rows represent \textcolor{myred}{social-interaction commonsense}, \textcolor{violet}{event-centered commonsense}, and \textcolor{darkgreen}{physical-entity commonsense}, respectively.}
\label{fig:ex_generations_events_physical}
\end{figure*}

\noindent
\textbf{Details about GPT2-XL Training.}
GPT2-XL \cite{radford2019language} is a transformer language model trained on 8 million webpages ($\sim$40G of text data). We finetune the language model on each commonsense knowledge graph by converting a tuple into a formatted text input -- e.g. \texttt{<head> <relation> [GEN] <tail> [SEP].} where [GEN] and [SEP] are special delimiter tokens that indicate the start and end of the tail of a given relation for a given head entity / event. At inference time, the head and relation of a tuple are given as input and the model's generation following the [GEN] token is recorded as its prediction of the tail entity. We finetuned GPT2-XL on each CSKG for one epoch, using a batch size of 32 and a learning rate of $5e-5$ on an Nvidia RTX-8000 GPU. The final trained models for each CSKG will be publicly released as part of our code release. In Figure~\ref{fig:ex_generations_events_physical}, we include a few examples of COMET(GPT2-XL) trained on \atomicTT.

\vspace*{2mm}
\noindent
\textbf{Details about BART Training.}
BART \citep{lewis-etal-2020-bart} is a denoising sequence-to-sequence pretrained language model.
Similar to previous transformer-based language models \cite{bert}, BART's pretraining objective is to recover its input, which is corrupted through various strategies such as token and span masking, and sentence permutation. 
For pretraining, BART uses a 160GB free-text dataset drawn from news, books, stories, and web texts.
We used the BART-large version of the model,\footnote{from HuggingFace's implementation~\cite{Wolf2019HuggingFacesTS}.} which has 24 layers, 1024-dimensional hidden states, 16 attention heads in its self-attention layers, and 406M total parameters.
We set the maximum length to be 24 and the minimum length to be 1.
For hyper-parameter search, we fine-tuned BART on each commonsense knowledge graph for one epoch with batch sizes \{64, 32, 16\}, learning rates \{1e-3, 1e-5, 1e-7\}, and three random seeds.

\vspace{2mm} \noindent \textbf{Details about GPT-3 Evaluation.}
We evaluate GPT-3 \cite{brown2020language} using OpenAI's language completion API. Similar to zero-shot evaluation on \gptxl, we use templates to evaluate the ability of the language model to generate a tail given the head and relation. We use the same templates as \gptxl. For priming examples, we prime each relation with 5 examples of heads and tails per relation, randomly selected from relations in the training set. We ran 3 random seeds to select priming examples to avoid spelling mistakes and other fragments from data collection. We ran with temperature 0.4.

\vspace{2mm} \noindent \textbf{Utility of Pre-trained Language Models.} 
As a means for establishing a control for the utility of pre-trained models in commonsense tasks, we trained an un-pretrained BART model and performed human evaluation for \atomicTT{}. We observe generation accuracy values of 54.9\% for Accept, 44.9\% for Reject, and 0.18\% for No Judgement, which is a significant drop in performance compared to the results for \cometbart{} in Table \ref{tab:human-eval-generations}. This indicates that pre-training does indeed provide a level of generalizations necessary for commonsense tasks.

\vspace*{2mm}
\noindent
\textbf{Additional Automated Evaluation.}
In order to have a direct comparison between automated and human evaluations, we report in Section \ref{sec:generalization} the automated metrics on the same test subsets that were used for human evaluation. For completeness, in this section, we provide the automated evaluation results on the full test sets (Table~\ref{tab:auto-eval-full-test}). These results confirm the findings of Section \ref{sec:generalization}.

\newpage
\begin{figure*}
\centering
\includegraphics[width=0.97    \textwidth]{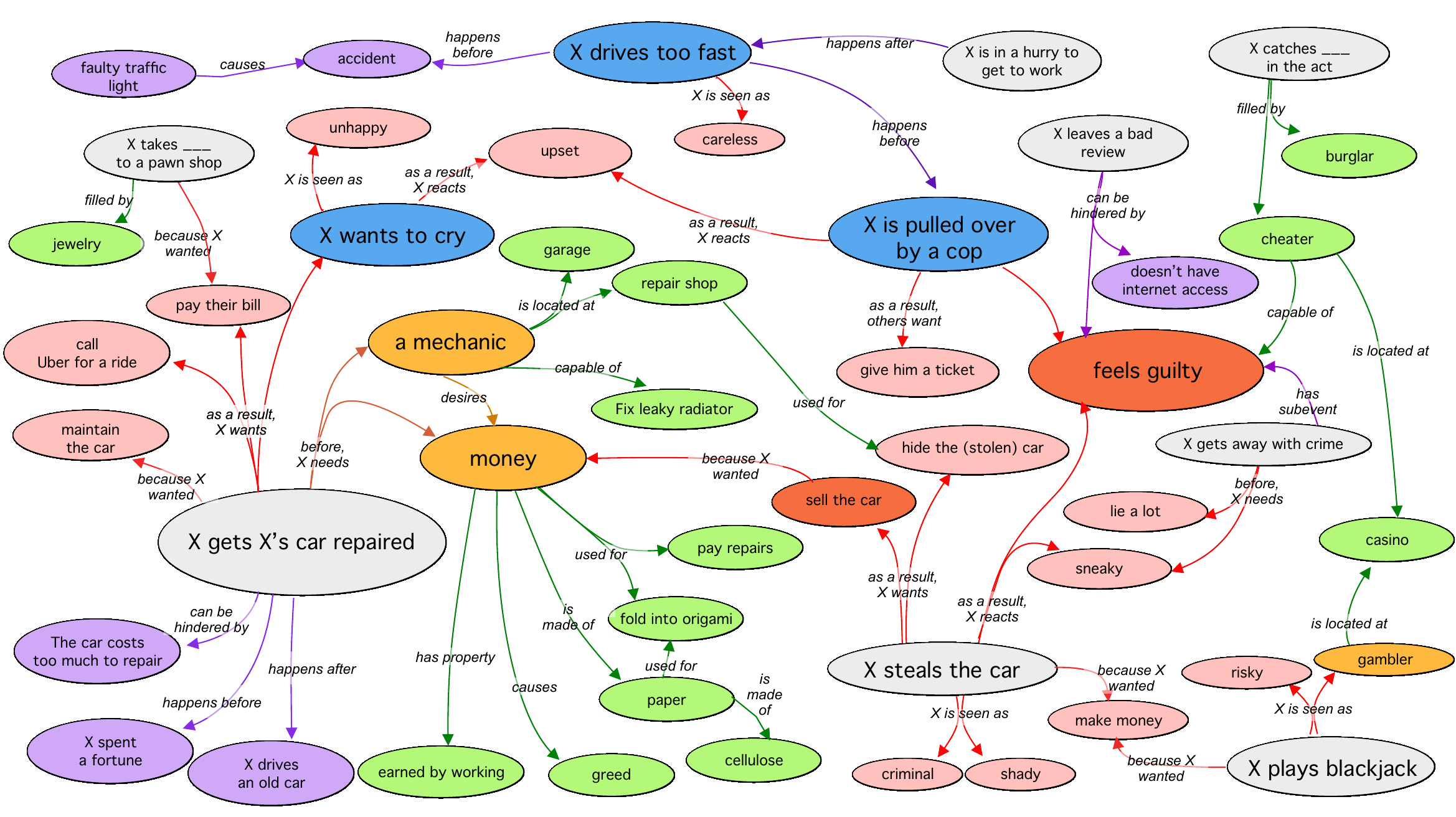}
\caption{A snapshot of commonsense knowledge relationships in \atomicTT{}. \textcolor{darkgray}{Gray} nodes represent events. \textcolor{myred}{Red}, \textcolor{violet}{purple} and \textcolor{darkgreen}{green} nodes represent \textcolor{myred}{social-interaction commonsense}, \textcolor{violet}{event-centered commonsense}, and \textcolor{darkgreen}{physical-entity commonsense}, respectively. Rest of the colors represent intersection of the categories.}
\end{figure*}

\input{sections/reproducibility_checklist}

%% file: tables/tab-split-sizes.tex
\begin{table}[ht]
\small
\setlength\tabcolsep{4pt} 
\begin{tabular}{lrrrr}
\toprule
\textbf{Knowledge Graph}            &   \multicolumn{1}{c}{\textbf{Train}} &   \multicolumn{1}{c}{\textbf{Dev}} &   \multicolumn{1}{c}{\textbf{Test}} &   \multicolumn{1}{c}{\textbf{All}} \\
\hline
\atomicTT &  1,076,880 &       102,024 &       152,209 &     1,331,113 \\
\atomic     &   709,993 &        79,599 &        87,480 &      877,072 \\
\conceptnet &   264,791 &         5,000 &        30,209 &      300,000 \\
\transomcs  &  5,424,478 &        10,243 &       100,033 &     5,534,754 \\
\hline
\end{tabular}
\caption{Number of tuples per KB and per split.}
\label{tab:split_sizes}
\end{table}

%% file: tables/tab-automated-metrics-full-test.tex
\begin{table*}[t]
  \centering
  \small
  \begin{tabular}{llcccccccc}
\toprule
                &       &   Bleu-1 &   Bleu-2 &   Bleu-3 &   Bleu-4 &   METEOR &   ROUGE-L &   CIDEr &   BERT Score \\
 \midrule
 \multirow{2}{*}{\atomicTT{}} & \cometgptxl{} &   0.401 &    0.247 &    0.168 &    0.123 &    0.288 &     0.473 &   0.620 &        0.632 \\
 & \cometbart{}       &   \bf 0.462 &    \bf 0.280 &    \bf 0.182 &    \bf 0.124 &    \bf 0.325 &    \bf 0.486 &   \bf 0.632 &        \bf 0.636 \\
  \midrule
 \multirow{2}{*}{\atomic{}} & \cometgptxl{}     &     0.429 &    0.300 &    0.225 &    \bf 0.187 &    0.297 &     0.527 &   0.754 &        0.638 \\
 & \cometbart{}            &     \bf 0.521 &    \bf 0.330 &    0.225 &    0.164 &    \bf 0.351 &    \bf  0.552 &  \bf  0.766 &     \bf    0.650 \\
  \midrule
 \multirow{2}{*}{\conceptnet{}} & \cometgptxl{} &    0.152 &   \bf  0.115 &  \bf   0.092 &  \bf   0.080 &  \bf   0.131 &   \bf   0.193 &  \bf  0.421 &      \bf   0.552 \\
 & \cometbart{}        &  \bf   0.169 &    0.108 &    0.069 &    0.046 &    0.127 &     0.180 &   0.350 &        0.532 \\
  \midrule
 \multirow{2}{*}{\transomcs{}} & \cometgptxl{} &     0.298 &    0.000 &    0.000 &    0.000 &    0.179 &     0.300 &   0.249 &        0.677 \\
 & \cometbart{}         &    \bf  0.351 &   \bf  0.216 &   \bf  0.004 &    0.000 &  \bf   0.201 &   \bf   0.352 & \bf   0.298 &      \bf   0.681 \\

 \bottomrule
\end{tabular}
\caption{Automated metrics for the quality of the \emph{tail} generations for the knowledge models \cometgptxl{} and \cometbart{}. Each approach uses greedy decoding for all test prefixes for each KG. Similar results were obtained on the 5K sampled prefixes that were randomly selected for the human evaluation (see Table \ref{tab:auto-eval}).}
\label{tab:auto-eval-full-test}
\end{table*}

 
 
 

%% file: sections/reproducibility_checklist.tex
\section{Additional Reproducibility Items}
\label{app:repro}

All experiments were conducted on a cluster with 8 GPUs of type NVIDIA Quadro RTX 8000 with 48 GB of GDDR6 memory each. To allow replication of results, whenever possible, a default, fixed value was assigned to the random seed that initializes the pseudo-random number generator, as specified in the source code. The details of the experimentation of the models (i.e. \gptxl{} and BART), including their hyper-parameter settings, are described in Appendix \ref{app:neural}. All the data as well as the source code required for conducting experiments will be made publicly available upon publication.